\documentclass[letterpaper]{article} %
\usepackage{aaai2026}  %
\usepackage{times}  %
\usepackage{helvet}  %
\usepackage{courier}  %
\usepackage[hyphens]{url}  %
\usepackage{graphicx} %
\usepackage{natbib}  %
\usepackage{caption} %
\usepackage{algorithm}
\usepackage{algpseudocode}
\usepackage{enumitem}
\usepackage{paralist}
\usepackage{newfloat}
\usepackage{listings}
\DeclareCaptionStyle{ruled}{labelfont=normalfont,labelsep=colon,strut=off} %
\floatstyle{ruled}
\newfloat{listing}{tb}{lst}{}
\floatname{listing}{Listing}
\newcommand{\supmat}{Appendix}

\usepackage{framed}
\newcounter{summary}
\newenvironment{summary}[1][]{\refstepcounter{summary}\par%
   \begin{framed}
   \noindent\textbf{Takeaway \thesummary.} \rmfamily
   } {
    \end{framed}
   }%

\nocopyright
   
\usepackage{booktabs}
\usepackage{multirow}
\usepackage{adjustbox}
\usepackage{siunitx}
\usepackage{amssymb}
\usepackage{nicefrac}
\usepackage{xspace}
\usepackage{listings}
\usepackage{nicefrac}
\usepackage{nicematrix}
\usepackage{xcolor}
\usepackage{comment}

\usepackage{cleveref}
\usepackage{csquotes}
\usepackage{subcaption}
\usepackage{tikz}
\usetikzlibrary{positioning, shapes, arrows.meta,calc}
\usepackage[T1]{fontenc}
\usepackage{aliascnt}

\DeclareMathOperator*{\argmax}{arg\,max}

\newcommand{\RR}{\mathbb{R}}
\newcommand{\size}[1]{\lvert #1 \rvert} %

\newcommand{\setX}{\mathcal{X}}
\newcommand{\setY}{\mathcal{Y}}

\newcommand{\adv}{\mathcal{A}} %
\newcommand{\dist}[1]{\lVert #1 \rVert} %

\newcommand{\advsuccrate}{\ensuremath{\mathsf{ASR}}\xspace}
\newcommand{\robustacc}{\ensuremath{\mathsf{RA}}\xspace}
\newcommand{\rat}{\ensuremath{\robustacc_\mathrm{T}}\xspace}
\newcommand{\raq}{\ensuremath{\robustacc_\mathrm{Q}}\xspace}

\newcommand{\mainacc}{\ensuremath{\mathsf{CA}}\xspace}

\usepackage{array}
\newcolumntype{H}{>{\setbox0=\hbox\bgroup}c<{\egroup}@{}}

\newcommand{\Batch}{\ensuremath{B}\xspace}

\newcommand{\Momen}{\ensuremath{\mu}\xspace}

\newcommand{\Lr}{\ensuremath{\eta}\xspace}

\newcommand{\Wd}{\ensuremath{\lambda}\xspace}

\makeatletter
\newcommand{\btriangle}{\mathpalette\btriangle@\relax}
\newcommand{\btriangle@}[2]{%
  \begingroup
  \sbox\z@{$\m@th#1\triangle$}%
  \makebox[\wd\z@]{%
    \raisebox{0.04\height}{%
      \resizebox{1.1\wd\z@}{0.96\ht\z@}{%
        $\m@th#1\blacktriangle$%
      }%
    }%
  }%
  \endgroup
}
\makeatother

\usepackage{dsfont}
\usepackage{amsthm}
\def\ddefloop#1{\ifx\ddefloop#1\else\ddef{#1}\expandafter\ddefloop\fi}
\def\ddef#1{\expandafter\def\csname bb#1\endcsname{\ensuremath{\mathbb{#1}}}}
\ddefloop ABCDEFGHIJKLMNOPQRSTUVWXYZ\ddefloop
\def\ddef#1{\expandafter\def\csname c#1\endcsname{\ensuremath{\mathcal{#1}}}}
\ddefloop ABCDEFGHIJKLMNOPQRSTUVWXYZ\ddefloop

\def\1{\mathds{I}}

\def\supp{\textup{supp}}

\newtheorem{proposition}{Proposition}

\theoremstyle{definition}
\newtheorem{definition}{Definition}
\theoremstyle{remark}

\theoremstyle{plain}

\newtheorem*{theorem*}{Theorem}

\renewcommand{\Pr}[1]{\mathrm{Pr}\left(#1\right)}

\usepackage{mathtools} %

\makeatletter
\newcommand*{\descbox}{\@ifstar{\let\@tempa\sp\desc@box}{\let\@tempa\sb\desc@box}}
\newcommand*{\desc@box}[2]{%
   {\mathop{\boxed{#2}}\limits\@tempa{{#1}}}%
}
\makeatother

\newcommand{\model}{{\mathcal{F}}}
\newcommand{\modelt}{{\mathcal{G}}}

\newcommand{\el}{{\ell_{\model}}}
\newcommand{\lt}{{\ell_{\modelt}}}

\theoremstyle{plain}

\title{Tuning for Two Adversaries: Enhancing the Robustness Against Transfer and Query-Based Attacks using Hyperparameter Tuning}
\author {
    Pascal Zimmer,
    Ghassan Karame
}
\affiliations {
    Ruhr University Bochum, Germany\\
    pascal.zimmer@rub.de, ghassan@karame.org
}

\begin{document}

\maketitle

\begin{abstract}
In this paper, we present the first detailed analysis of how training hyperparameters---such as learning rate, weight decay, momentum, and batch size---influence robustness against both transfer-based and query-based attacks.

Supported by theory and experiments, our study spans a variety of practical deployment settings, including centralized training, ensemble learning, and distributed training. We uncover a striking dichotomy: for transfer-based attacks, decreasing the learning rate significantly enhances robustness by up to $64\%$. In contrast, for query-based attacks, increasing the learning rate consistently leads to improved robustness by up to $28\%$ across various settings and data distributions. Leveraging these findings, we explore---for the first time---the training  hyperparameter space to jointly enhance robustness against both transfer-based and query-based attacks. Our results reveal that distributed models benefit the most from hyperparameter tuning, achieving a remarkable tradeoff by simultaneously mitigating both attack types more effectively than other training setups.
\end{abstract}

\begin{links}
    \link{Code}{https://github.com/RUB-InfSec/tuning_for_two_adversaries}
\end{links}

\section{Introduction}

Despite their growing popularity, machine learning systems remain vulnerable to relatively simple manipulations of their inputs~\cite{DBLP:journals/corr/SzegedyZSBEGF13,DBLP:conf/pkdd/BiggioCMNSLGR13}. In particular, adversarial examples pose a significant threat to the application of deep neural networks (DNNs) in safety-critical domains, including autonomous driving and facial recognition. 
Recent strategies, known as black-box attacks~\cite{chenHopSkipJumpAttackQueryEfficientDecisionBased2020, 10.1007/978-3-030-58592-1_29, chenRethinkingModelEnsemble2024}, adopt a practical and realistic threat model in which the attacker lacks access to the classifier’s internal details and training data and can either (1) build a surrogate model to generate adversarial examples that transfer to the target (called transfer-based attacks) in one-shot attacks, or (2) interact with the model through oracle access---by observing its outputs in response to specific inputs (aka. query-based attacks) as is the case in machine-learning-as-a-service (MLaaS) settings. 

A range of defense strategies has been proposed to counter adversarial examples, with adversarial training~\cite{goodfellowExplainingHarnessingAdversarial2015} still regarded as the gold standard among training-time defenses. However, this method is computationally expensive, as it requires retraining the target model using adversarially perturbed \emph{pre-generated} inputs. To reduce overhead, lightweight test-time defenses---such as input transformations~\cite{DBLP:conf/iclr/GuoRCM18}---have been explored. However, these methods frequently compromise performance on clean inputs when aiming for strong robustness.
Moreover, the rapid evolution of machine learning research has led to defenses being developed in isolation for either transfer-based or query-based attacks. This raises significant concerns about their ability to effectively enhance robustness across the full spectrum of black-box attack scenarios.

Given the lack of bullet-proof solutions to thwart adversarial examples, a natural question is whether the training hyperparameters themselves present an opportunity in this setting, as they are inherently tied to the target model and are independent of specific attacks. While prior studies have shown that factors such as the scheduler, optimizer, and architecture have only minimal impact on robustness~\cite{DBLP:conf/aaai/AndreinaZK25}, hyperparameters---such as learning rate, momentum, weight decay, and batch size---significantly influence model smoothness due to their regularizing effects, a property closely associated with robustness to adversarial attacks~\cite{moosavi-dezfooliRobustnessCurvatureRegularization2019, zhangWhyDoesLittle2024, demontis2019adversarial}. 
Unfortunately, to the best of our knowledge, no prior work has conducted a precise analysis of how hyperparameter tuning affects robustness against specific attacks in the black-box setting. 

In this paper, we address this gap and present the first precise analysis of how training hyperparameters—such as learning rate, weight decay, momentum, and batch size—influence robustness against strong transfer-based and query-based black-box attacks. Our analysis spans a range of popular machine learning deployment scenarios, including centralized setups (with a single training and inference instance), ensemble learning, and distributed learning, where both training and inference are distributed across multiple nodes. We also account for different data distribution settings, encompassing both independent and identically distributed (i.i.d.) and non-i.i.d. training data.
More specifically, we aim to answer the following research questions:

\begin{description} 
\item[\textbf{RQ 1}] To what extent do training hyperparameters influence robustness against transfer-based attacks?
\item[\textbf{RQ 2}] Similarly, how do training hyperparameters influence robustness against query-based attacks? 
\item[\textbf{RQ 3}] Is there an instantiation that naturally lends itself to effective tuning against both transfer-based and query-based attacks?
\end{description}
Supported by theory and extensive experiments on CIFAR-10 and ImageNet, our results reveal a striking contrast. On the one hand, we find that decreasing the learning rate can significantly enhance robustness against transfer-based attacks---by up to 64\%---with this improvement consistently holding across different ML deployment scenarios, including centralized, ensemble, and distributed setups.
Surprisingly, we observe the opposite trend for query-based attacks: increasing the learning rate leads to a robustness gain of up to 28\% across all considered deployments and data distributions. 
Last but not least, we explore how to effectively balance hyperparameter tuning to navigate the trade-offs between robustness to transfer-based and query-based attacks. Through an extensive search using the NSGA-II algorithm, we demonstrate that there exist well-chosen hyperparameter configurations that strike a strong trade-off, enhancing robustness against both transfer-based and query-based attacks, \emph{when compared to state-of-the-art defenses}, such as adversarial training~\cite{rebuffiFixingDataAugmentation2021, gowalUncoveringLimitsAdversarial2021} and JPEG compression~\cite{DBLP:conf/iclr/GuoRCM18}.

\section{Background \& Related Work}
\label{sec:background}

\subsection{Black-Box Attacks}
\noindent \textbf{Transfer-based Attacks.} 
In transfer attacks, adversaries train surrogate models and apply white-box attacks on them. Success depends on the similarity between the surrogate and target models.
To improve transferability in black-box settings, various techniques have emerged, including momentum~\cite{DBLP:conf/cvpr/DongLPS0HL18}, input diversification~\cite{DBLP:conf/cvpr/WuSLK21}, model ensembling, and sharpness-aware minimization~\cite{chenRethinkingModelEnsemble2024}. Notably, \cite{chenRethinkingModelEnsemble2024} showed that reducing sharpness and aligning gradients boosts transferability. Other studies explore how model architecture, capacity~\cite{demontis2019adversarial}, and regularization-based control of smoothness and gradients~\cite{zhangWhyDoesLittle2024} affect transferability bounds.

\noindent \textbf{Query-based Attacks.} 
Query-based attacks interact directly with the target model. Here, attackers are limited by a query budget $Q$ and the granularity of accessible information. Score-based attacks~\cite{chenZOOZerothOrder2017, tuAutoZOOMAutoencoderBasedZeroth2019, 10.1007/978-3-030-58592-1_29, ilyasBlackboxAdversarialAttacksa} use model output scores or probabilities, while decision-based attacks~\cite{chenHopSkipJumpAttackQueryEfficientDecisionBased2020} rely only on the top-1 predicted label.
Most attacks begin with locating the decision boundary via binary search, then use gradient estimation~\cite{chenZOOZerothOrder2017} or geometric strategies to approach the boundary while maintaining adversarial properties. SquareAttack~\cite{10.1007/978-3-030-58592-1_29}, a state-of-the-art score-based method, employs square-shaped perturbations and a simple random search scheme.

\subsection{Hyperparameter Tuning}
Despite the theoretical insights, hyperparameter tuning in practice remains largely ad hoc with limited insights on how hyperparameter configurations influence black-box robustness~\cite{bagdasarianMithridatesAuditingBoosting2024}.

\noindent \textbf{Implicit Regularization.} 
Training hyperparameters implicitly regularize neural networks by influencing the geometry of the learned solutions, often characterized by their flatness or sharpness---quantified via the largest eigenvalue of the Hessian~\cite{kaurMaximumHessianEigenvalue}. In full-batch gradient descent, the sharpness of the converged solution scales inversely with the learning rate $\eta$~\cite{cohenGradientDescentNeural2022}. This inverse relationship approximately holds for small-batch stochastic gradient descent (SGD) as well~\cite{kaurMaximumHessianEigenvalue}. Other hyperparameters such as batch size $B$ and weight decay $\lambda$ also affect sharpness: increasing the ratio $\eta/B$ tends to decrease the largest Hessian eigenvalue~\cite{jastrzebskiThreeFactorsInfluencing2018}, while the product $\eta\lambda$ has been linked to an implicit form of Jacobian regularization~\cite{dangeloWhyWeNeed}.

\noindent \textbf{From Parameter Space to Input Gradients.} 
On the other hand, several studies identified a form of gradient pressure originating from the parameter space to the input space~\cite{maLinearStabilitySGD,dherinWhyNeuralNetworksa}, indicating that flatter solutions tend to yield smoother decision boundaries.

\subsection{Defenses against Black-Box Attacks}
\noindent \textbf{Adversarial training.}
While many defenses attempt to mitigate model evasion attacks with lightweight test-time input transformations~\cite{DBLP:conf/iclr/GuoRCM18} or auxiliary models~\cite{DBLP:conf/icml/NieGHXVA22}, adversarial training remains the de facto standard. The basic idea is to augment the training data with adversarial examples or to use a fine-tuning phase, which adds a considerable computational complexity to this training-time defense. Here, it is crucial to ensure that the models do not overfit to the type of adversarial noise found in their training samples, but generalize to other (unseen) attacks~\cite{DBLP:conf/iclr/TramerKPGBM18}. Techniques to further boost performance of adversarial training can, among others, leverage data augmentation~\cite{rebuffiFixingDataAugmentation2021,gowalUncoveringLimitsAdversarial2021} or topology alignment~\cite{kuangDefenseAdversarialAttacks2024}. 

\noindent \textbf{Diverse ensemble strategies.}
A large body of works aims at improving the robustness of model ensembles as a whole, i.e., by ensuring that each ensemble member is dissimilar to the other members, making sure that an adversarial example only transfers to a few members~\cite{yangTRSTransferabilityReduced, yangDVERGEDiversifyingVulnerabilities, pangImprovingAdversarialRobustness, dengUnderstandingImprovingEnsemble, caiEnsembleinOneEnsembleLearning2023, kariyappaImprovingAdversarialRobustness2019}, e.g., by incorporating adversarial training with other ensemble members or by ensuring pairwise negative/orthogonal gradients.

\section{Methodology}
\label{sec:method}

\subsection{Preliminaries and Definitions}
We denote the sample and label spaces with $\mathcal{X}$ and $\mathcal{Y}$, respectively, and the training data with $\mathcal{S}=(x_i,y_i)^M_{i=1}$, with $x_i \in \mathbb{R}^d, y_i \in \mathbb{R}^c$ and training set size $M$. We assume that $\mathcal{S}$ is sampled from the underlying data distribution $\mathcal{D}$.

A DNN-based classifier $\cF_{\theta}: \mathcal{X}\to[0,1]^{c}$ is a function (parameterized by $\theta$) that, given an input $x$, outputs the probability that the input is classified as each of the $n=|\mathcal{Y}|$ 
classes. The prediction of the classifier can be derived as $y=C(x):=\argmax_{i \in [n]}({\cF_{\theta}^i}(x))$.

\vspace{0.25 em}\noindent\textbf{Adversarial examples.} We define an adversarial example $x'$ as a genuine image $x$ to which carefully crafted adversarial noise is added, i.e., $x' = x + \delta$ for a small perturbation~$\delta$ such that~$x'$ and~$x$ are perceptually indistinguishable to the human eye and yet are classified differently.
Given a genuine input~$x_0\in \RR^d$ predicted as~${C(x_0) = s}$ (source class), $x'$ is an \emph{adversarial example} of~$x_0$ if 
$C(x')~\neq~s$ and
$\dist{x'-x_0}_p \leq \varepsilon$ for a given distortion bound~$\varepsilon\in \RR^+$ and $l_p$ norm. 
The attacker searches for adversarial inputs~$x'$ with low distortion while maximizing a loss function $\cL_\model$ ($\theta$ omitted for clarity), e.g., cross-entropy loss. Formally, the optimization problem is defined by: $
    \delta = \argmax_{\lVert\delta\rVert_p\leq\varepsilon}\cL_\model(x+\delta,y)$.

\begin{definition}[Stochastic Gradient Descent]%
    SGD is a (noisy) optimization procedure for minimizing a loss function in training neural networks and is often combined with momentum and weight decay. For a minibatch of size $B$ and weight decay $\lambda$, we define the gradient at iteration $t$ as:
    \begin{align}
    \label{eq:grad_est}
        g_t &= \textstyle\frac{1}{B}\sum_{i=1}^{B}\nabla_\theta\cL(\cF_\theta(x_{i_t}), y_{i_t}) + \lambda{\theta_t},
    \end{align}
    while drawing indices i.i.d. out of the uniform distribution $i_t\sim\mathbb{U}([N])$.
    The current velocity $v$ weighs the velocity of the last iteration with momentum $\mu$. The model parameters are then updated with learning rate $\eta$ as follows:
    \begin{align}   
    \label{eq:theta_update}
        v_t = \mu v_{t-1} + g_t, \hspace{0.2 cm}
        \theta_{t+1}=\theta_t-{\eta}v_t.
    \end{align} 
\end{definition}

\begin{definition}[Model Smoothness] 
\label{def:smoothness}
    Given a model $\model$ and a data distribution $\cD$, the upper smoothness of $\model$ on $\cD$~\cite{zhangWhyDoesLittle2024} is defined as:
    \begin{equation}
    \overline{\sigma}_\model=\textstyle\sup_{(x,y)\sim\cD}\sigma(\nabla_x^2\cL_\model(x,y)),
\end{equation}
    where $\sigma(\cdot)$ denotes the largest eigenvalue, and $\nabla_x^2\cL_\model(x,y)$ the Hessian matrix computed w.r.t. $x$. 
\end{definition}

\begin{definition}[Gradient Similarity]
\label{def:gradient_sim}
    Given two models $\model$ and $\modelt$ and their respective loss functions $\cL_\model$ and $\cL_\modelt$, we can compute the gradient similarity~\cite{demontis2019adversarial,zhangWhyDoesLittle2024} based on the cosine similarity as follows:
    \begin{equation}\label{eq:s}
    \cS(\cL_\model, \cL_\modelt, x,y) =\frac{\nabla_x{\cL_\model(x, y)}^\intercal\nabla_x{\cL_\modelt(x,y)}}{\lVert\nabla_x{\cL_\model(x,y)}\rVert_2\lVert\nabla_x{\cL_\modelt(x,y)}\rVert_2}
    \end{equation}
    and the upper gradient similarity based on the supremum as: 
    \begin{equation}
        \overline{\cS}(\cL_\model, \cL_\modelt)=\textstyle\sup_{(x,y)\sim\cD}\cS(\cL_\model, \cL_\modelt, x, y)
    \end{equation}
\end{definition}

\subsection{Main Intuition}
Hyperparameter selection is critical for training well-performing models and is a core component of any machine learning pipeline. Since hyperparameters induce implicit regularization, we investigate their impact on model robustness against evasion attacks in a black-box threat model.

\begin{figure}[tb]
    \centering
    \includegraphics[width=\linewidth]{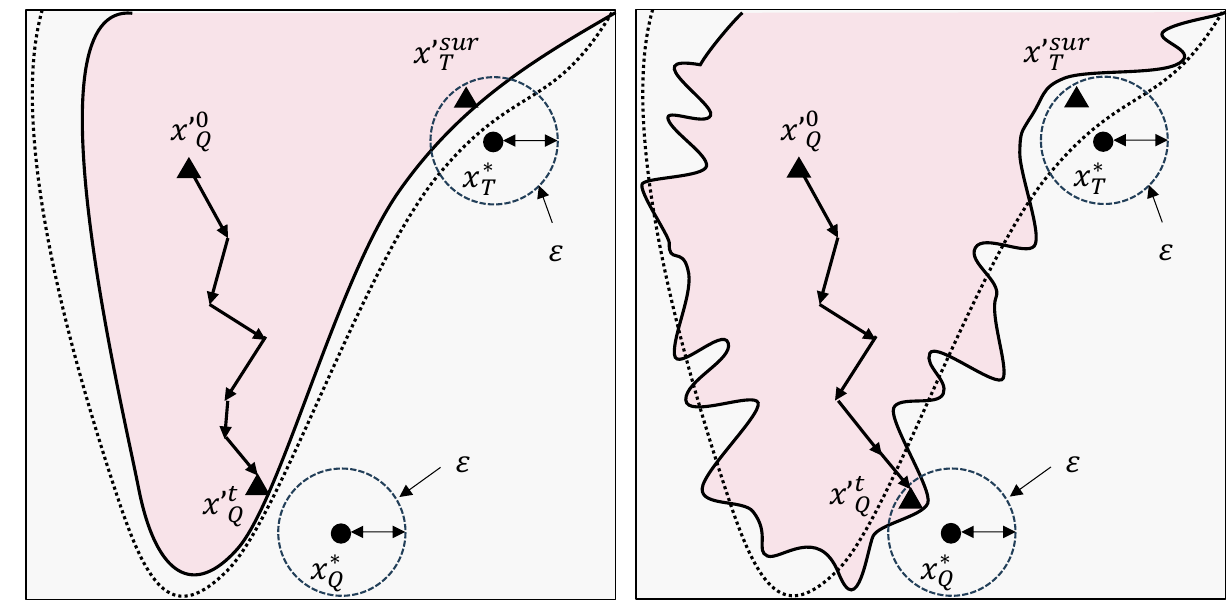}
    \caption{Tension between adversarial examples ($\btriangle$) for transfer-based ($x'_T$) and query-based attacks ($x'_Q$) on a smooth (left) and a less smooth (right) model. The solid line and dotted line represent the decision boundaries for the target and surrogate model, respectively. The dashed line represents the $\varepsilon$ constraint around a benign sample ($x^*_{T/Q}$ / \tikz\draw[black,fill=black] (0,0) circle (.7ex);).}
    \label{fig:tension}
\end{figure}

In this work, we demonstrate that: (1) less smooth models are more robust to transfer-based attacks, as adversarial examples crafted on smoother surrogates are less likely to transfer to models with more variable loss landscapes~\cite{demontis2019adversarial} (cf.~\Cref{prop:transfer}); and (2) smoother models exhibit greater robustness against query-based attacks~\cite{moosavi-dezfooliRobustnessCurvatureRegularization2019}. In~\Cref{fig:tension} (left), we see that adversarial examples generated towards a (typically smooth) surrogate model ($x'^{sur}_{T}$) transfer well to a smooth target model. However, the target remains robust to query-based attacks, as no adversarial example with perturbation magnitude $|\varepsilon|$ can be found within the query budget. In contrast, \Cref{fig:tension} (right) shows that a less smooth model hinders transferability due to gradient misalignment but is more vulnerable to query-based attacks, which converge more easily.

\begin{figure}[tb]
    \centering
    \includegraphics[width=.245\textwidth]{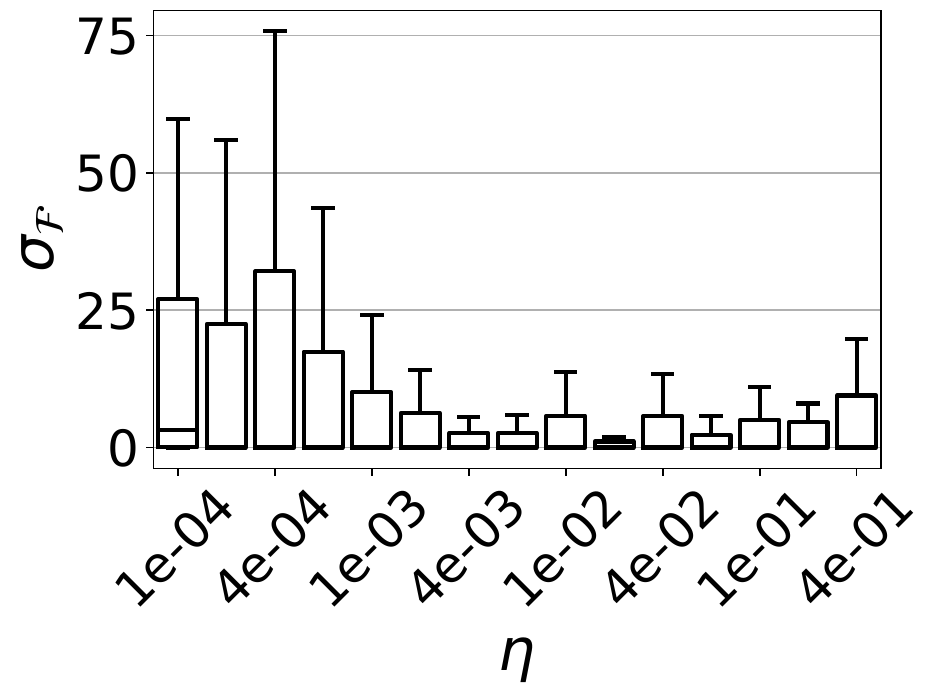}%
\includegraphics[height=.385\linewidth]{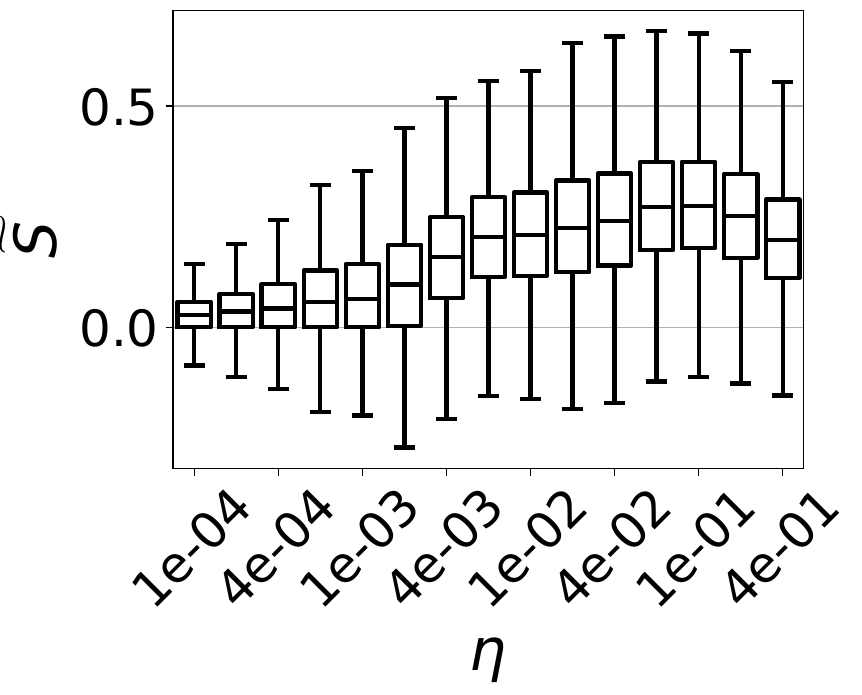}
    \caption{Impact of sharpness in parameter space (due to change of the learning rate hyperparameter) and smoothness in input space (left) and gradient similarity (right).}
    \label{fig:theory_exps}
\end{figure}

Adapting the results from ~\cite{cohenGradientDescentNeural2022,kaurMaximumHessianEigenvalue,jastrzebskiThreeFactorsInfluencing2018,dangeloWhyWeNeed,maLinearStabilitySGD,dherinWhyNeuralNetworksa}, 
we relate the largest eigenvalue of the loss Hessian in parameter space to that in input space and obtain:
\begin{equation}
\label{eq:theta_x}
    \sigma(\nabla_\theta^2\cL_\model(x, y)) \sim \sigma(\nabla_x^2\cL_\model(x, y))
\end{equation}
We empirically verify this connection in~\Cref{fig:theory_exps} (left) and provide more details in the \supmat. 

\vspace{0.25 em}\noindent\textbf{1) Transferability \& Model Smoothness.}
Let $T_r$ denote the transferability for a benign sample $x$ and an adversarial example $x'$ which is generated on a surrogate model $\model$, but evaluated on a target model $\modelt$, as follows:
$T_r(\cF, \cG, x) = \mathds{1} [\cF(x)=y \wedge \cG(x)= y \wedge \cF(x')\neq y \wedge \cG(x')\neq y]$.

\begin{proposition}[Less smooth models exhibit high robustness against transfer-based attacks]
\label{prop:transfer}
    For a surrogate model $\model$ (with smoothness $\overline{\sigma}_\model$) and target model $\modelt$ (with smoothness $\overline{\sigma}_\modelt$), the upper bound on transferability $\Pr{(T_r(\cF, \cG, x)}~=~1)$ decreases when $\modelt$'s smoothness also decreases (i.e., {$\overline{\sigma}_\modelt$} increases).
\end{proposition}
In the following, we first determine smoothness and gradient similarity to be the main contributors to transferability, and then focus on how the regularization properties of hyperparameters reduce smoothness and gradient similarity and hence reduce the upper bound on transferability. 

Based on~\cite{yangTRSTransferabilityReduced,zhangWhyDoesLittle2024}, we assume models $\cF$ and $\cG$ are $\overline{\sigma}_\model$-smooth and $\overline{\sigma}_\modelt$-smooth, respectively, with bounded gradient magnitude, i.e., $\|\nabla_x \cL_{\cF}(x,y)\| \le B_\model$ and $\|\nabla_x \cL_{\cG}(x,y)\| \le B_\modelt$ for any $x \in \cX$, $y \in \cY$.
            We consider an untargeted attack with perturbation ball $\|\delta\|_2 \le \varepsilon$.
            When the attack radius $\varepsilon$ is small such that the denominator is larger than 0, 
            the transferability can be upper bounded by:
            \resizebox{0.485\textwidth}{!}{
            \begin{minipage}{\linewidth}
            \begin{align*}
                    &\Pr{T_r(\cF, \cG, x) = 1} 
                    \\
                    &\le
                    \dfrac{\xi_\model }{\displaystyle \min_{\substack{x\in\cX, y'\in\cY:\\ (x,y) \in \supp(\cD),\\y' \neq y}} \, \cL_\cF(x, y') - \varepsilon B_\model \left(1 + \sqrt{\frac{1+\overline{\cS}(\cL_\model,\cL_\modelt)}{2}}\right) - \overline{\sigma}_\model\varepsilon^2}\\
                    &+
                    \dfrac{\xi_\modelt}{\displaystyle \min_{\substack{x\in\cX, y'\in\cY:\\ (x,y) \in \supp(\cD),\\y' \neq y}} \, \cL_\cG(x, y') - \varepsilon B_\modelt \left(1 + \sqrt{\frac{1+\overline{\cS}(\cL_\model,\cL_\modelt)}{2}}\right) - \overline{\sigma}_\modelt\varepsilon^2}
            \end{align*}
            \end{minipage}
            }
            Here $\xi_\cF$ and $\xi_\cG$ are the \emph{empirical risks} of models $\cF$ and $\cG$, respectively, defined relative to a differentiable loss.
            The $\supp(\cD)$ is the support of benign data distribution, i.e., $x$ is the benign data and $y$ is its associated true label. The full proof can be found in the \supmat.
            \label{trans_upper_bound}

We observe that the surrogate model $\model$ and target model $\modelt$ equally contribute to this upper bound, with smoothness and gradient similarity being the main contributing properties.\footnote{This contrasts with prior work~\cite{yangTRSTransferabilityReduced} that assumes that smoothness and gradient magnitude are identical for $\model$ and $\modelt$.}
Naturally, our aim is to reduce this upper bound to reduce  the transferability of the model. 

While the smoothness $\overline{\sigma}_\modelt$ (cf.~\Cref{def:smoothness}) (related to $B_\modelt$) is specific to the target, the gradient similarity $\overline{\cS}(\cL_\model, \cL_\modelt)$ (cf.~\Cref{def:gradient_sim}) is a shared quantity with $\model$ and \emph{positively correlated to the upper bound}. The relationship of regularization and alignment of input space gradients, i.e., their similarity, has been investigated by~\cite{zhangWhyDoesLittle2024} and found to be positively correlated, with varying impact depending on the used regularizer (which we empirically verify in~\Cref{fig:theory_exps} (right) with more details in the \supmat). As a result, the regularization properties of hyperparameters  reduce smoothness and gradient similarity between the target and surrogate models, which in turn decreases the upper bound on transferability---thereby concluding this proposition. 

\vspace{0.25 em}\noindent\textbf{2) Robustness against Query Attacks \& Model Smoothness.}
The gradient of the loss surface in the input space, estimated by query-based attacks, is typically bounded by the true model gradient. More specifically, smoother models typically improve robustness by requiring a larger perturbation $\delta$ to successfully evade the model.

Namely, given a model $\model$, input $x$ and label $y$, let $g~=~\nabla_x\cL_\model(x, y)$ and $c:=t-\cL_\model(x, y) \geq 0$ with $t$ as loss threshold. Further, we consider ${\sigma}_\model \geq 0$ as the largest eigenvalue of the Hessian matrix $\nabla_x^2\cL_\model(x, y)$, and $u_d$ as the respective eigenvector. For a perturbation $\delta$, we now have the following bounds on the perturbation and hence robustness of a model as shown by~\cite{moosavi-dezfooliRobustnessCurvatureRegularization2019}:
\begin{align}
\label{eq:query}
    \textstyle\frac{c}{\dist{g}}-2\sigma_\model\frac{c^2}{\dist{g}^3}\leq\dist{\delta}\leq\frac{c}{g^\intercal u_d}
\end{align}
Here, we see that smooth models with bounded (input) gradients exhibit higher robustness than less smooth ones. Concretely, increasing the curvature $\sigma_\model$, reduces the norm of $\delta$, illustrating that smaller perturbations are sufficient for a successful adversarial example, while smaller curvature increases the required perturbation. 

\vspace{0.25 em}\noindent \textbf{Interpretation.}
Based on~\Cref{prop:transfer} and our observations, we notice an inherent tension between choice of hyperparameters and robustness against transfer- and query-based attacks. Concretely, smoothness is beneficial for the robustness against query-based attacks (cf. Equation~\eqref{eq:query}) but detrimental for the robustness against transfer-based attacks (cf.~\Cref{prop:transfer}), as this enables adversarial examples to transfer better from the typically smooth surrogates. The converse also holds: robustness against transfer-based attacks is improved with less smooth models and a decrease in gradient similarity, but weakened against query-based attacks by lowering the required perturbation magnitude.

Note that while the attacker has full control over the surrogate model $\model$, the target model $\modelt$ can be chosen arbitrarily by the defender. \emph{We therefore propose using hyperparameter-induced implicit regularization to adjust model smoothness, enabling a trade-off between robustness to transfer-based and query-based attacks.}

\section{Experimental Setup}
\label{sec:exps}

\noindent\textbf{Setup.}
All our experiments are run on an Ubuntu 24.04 machine with two NVIDIA A40 GPUs, two AMD EPYC 9554 64-core processors, and 512 GB of RAM. We used Python 3.11.11, CUDA 12.5, Lightning 2.5.0, and Ray Tune 2.40.0.

\vspace{0.25 em}\noindent\textbf{Attack Selection.}
We evaluate our approach against (i) a state-of-the-art transfer-attack, the Common Weakness attack~\cite{chenRethinkingModelEnsemble2024}, which uses an ensemble of surrogate models, and leverages sharpness aware minimization (SAM) to find solutions on a smooth area of a high loss region, and against (ii) a state-of-the-art query-based attack, the SquareAttack~\cite{10.1007/978-3-030-58592-1_29} from AutoAttack~\cite{DBLP:conf/icml/Croce020a}.

We use a perturbation budget of $\varepsilon=8/255$ under an $l_\infty$ norm and generate adversarial examples for $1000$ randomly sampled images from the test set for the transfer- and query-based attacks, respectively. For the transfer-based attacks, we consider an ensemble of 10 models trained with default parameters on various architectures. For the query-based attack, we assume a query budget of $Q=500$.
\begin{table}
    \centering
    \begin{adjustbox}{max width=0.95\linewidth}
    \begin{tabular}{lcll}
        \toprule
        \textbf{Parameter} & \textbf{Variable} & \textbf{Default} & \textbf{Range} \\\midrule
        Learning rate &\Lr & 0.1 &$[0.0001, 0.4]$ \\
        \midrule        
        Weight decay & \Wd & 0.0005 &$[0.000001, 0.01]$ \\
        \midrule
        Momentum & \Momen& 0.9 &$[0.8, 0.99]$\\
        \midrule
         \multirow{1}{*}{\shortstack[l]{Batch size}} &\multirow{1}{*}{\Batch} & 128 &\multirow{1}{*}{$[32, 2048]$}\\
        \bottomrule
    \end{tabular}
    \end{adjustbox}
    \caption{Typical SGD hyperparameters $\cH$ for training a CIFAR-10 model and their value ranges in our experiments.}
    \label{tab:parameters}
\end{table}

\vspace{0.25 em}\noindent\textbf{Datasets \& Models.}
We evaluate our approach on datasets of varying input dimensions and number of classes, i.e., CIFAR-10~\cite{cifar10} and ImageNet~\cite{russakovskyImageNetLargeScale2015} datasets. The former contains $50,000$ train and $10,000$ test images of size $32 \times 32$ pixels, divided into 10 classes. The latter contains $1.2$ million training images and a validation set of $50,000$ images. For both datasets, we focus on models of the ResNet family~\cite{heDeepResidualLearning2016} due to their wide adoption. We include additional results on MobileNetV2 in the \supmat. 

In addition to a basic centralized model training on the full dataset, we also consider various ensemble instantiations that use logit averaging (before softmax) for inference:
\begin{compactitem}
    \item Deep Ensemble (full dataset, different initialization)
    \item Distributed IID Ensemble (disjoint dataset with i.i.d. data)
    \item Distributed Non-IID Ensemble (disjoint dataset with non-i.i.d. data using Dirichlet distribution with $\alpha=0.9$)
\end{compactitem}

\begin{figure*}[tb]
    \centering
    \begin{subfigure}[t]{.245\textwidth}
        \centering
        \includegraphics[width=\linewidth]{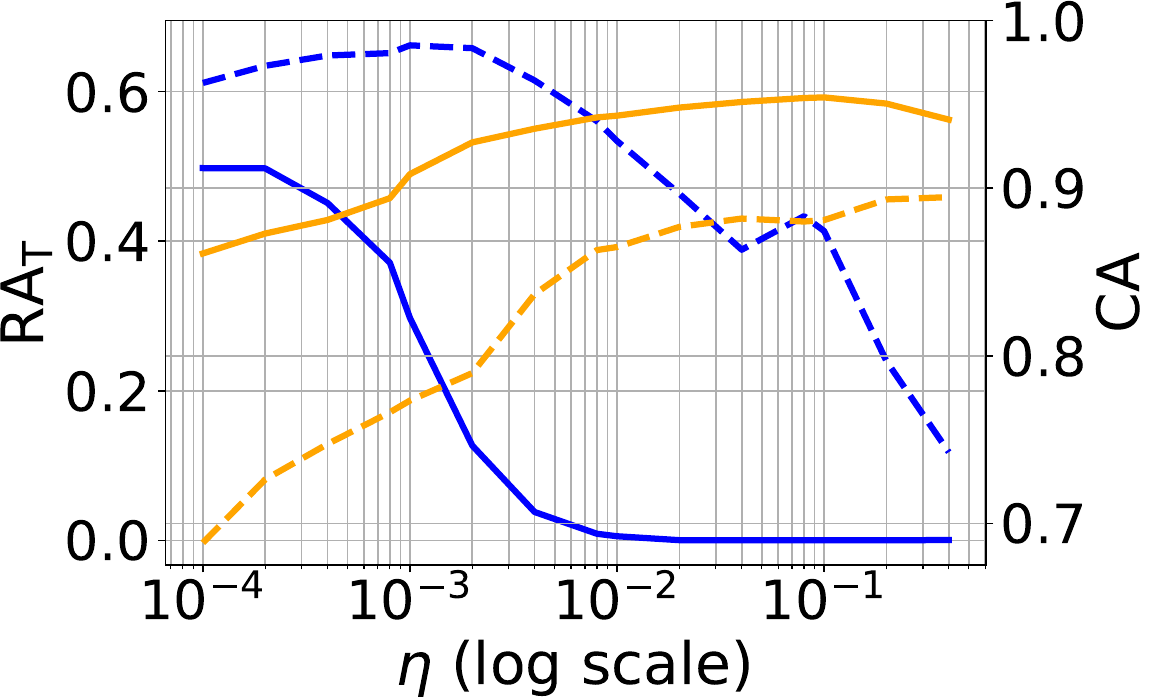}
        \caption{$\cA_T$ with varying learning rate}
   	    \label{fig:transfer_ablation_lr}
    \end{subfigure}
        \begin{subfigure}[t]{.245\textwidth}
        \centering
\includegraphics[width=\linewidth]{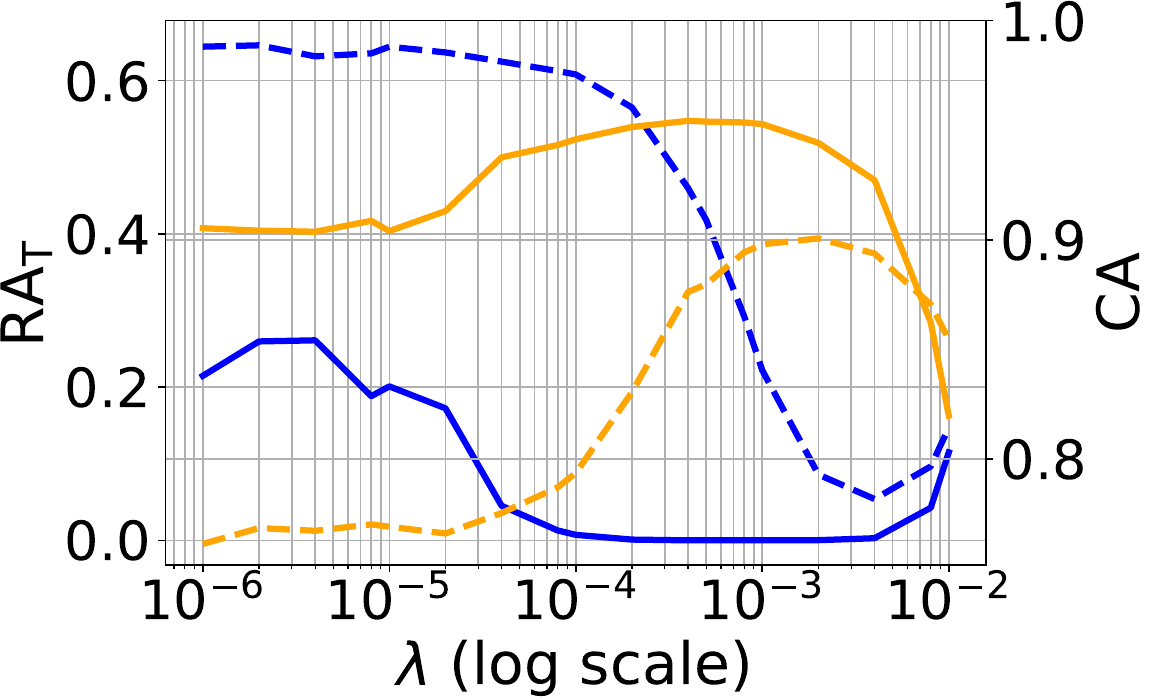}\caption{$\cA_T$ with varying weight decay}  
\label{fig:transfer_ablation_wd}
\end{subfigure}
        \begin{subfigure}[t]{.245\textwidth}
        \centering
\includegraphics[width=\linewidth]{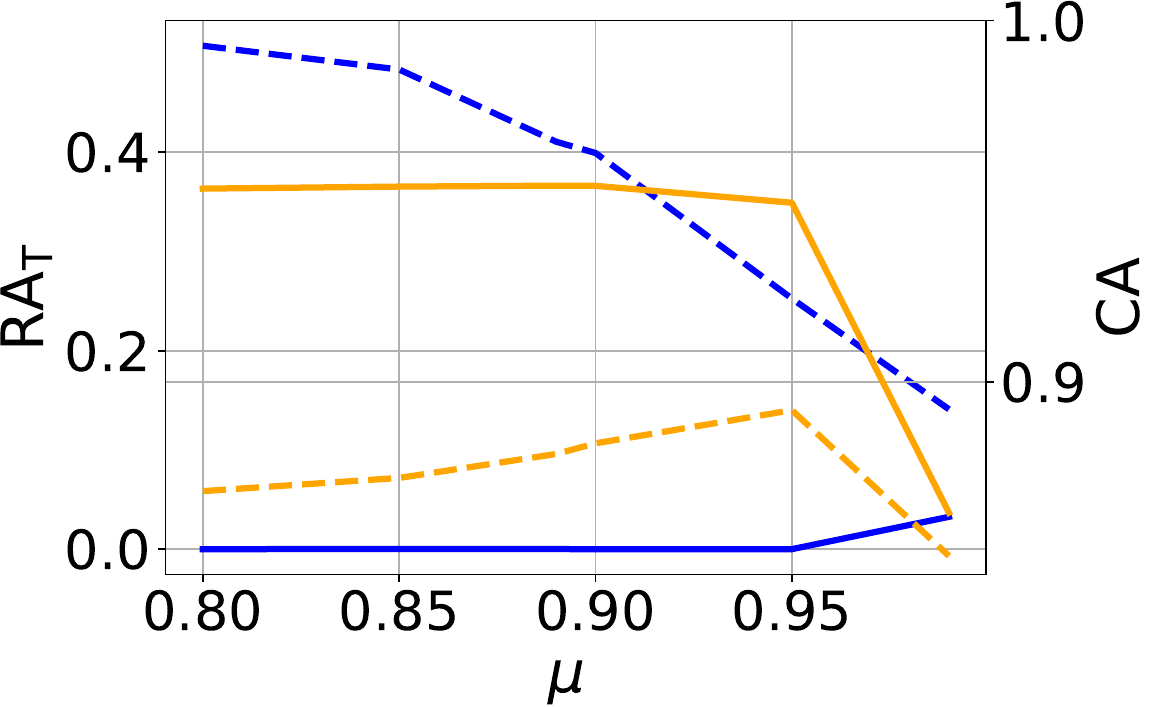}\caption{$\cA_T$ with varying momentum} 
\label{fig:transfer_ablation_mom}
\end{subfigure}
         \begin{subfigure}[t]{.245\textwidth}
        \centering
\includegraphics[width=\linewidth]{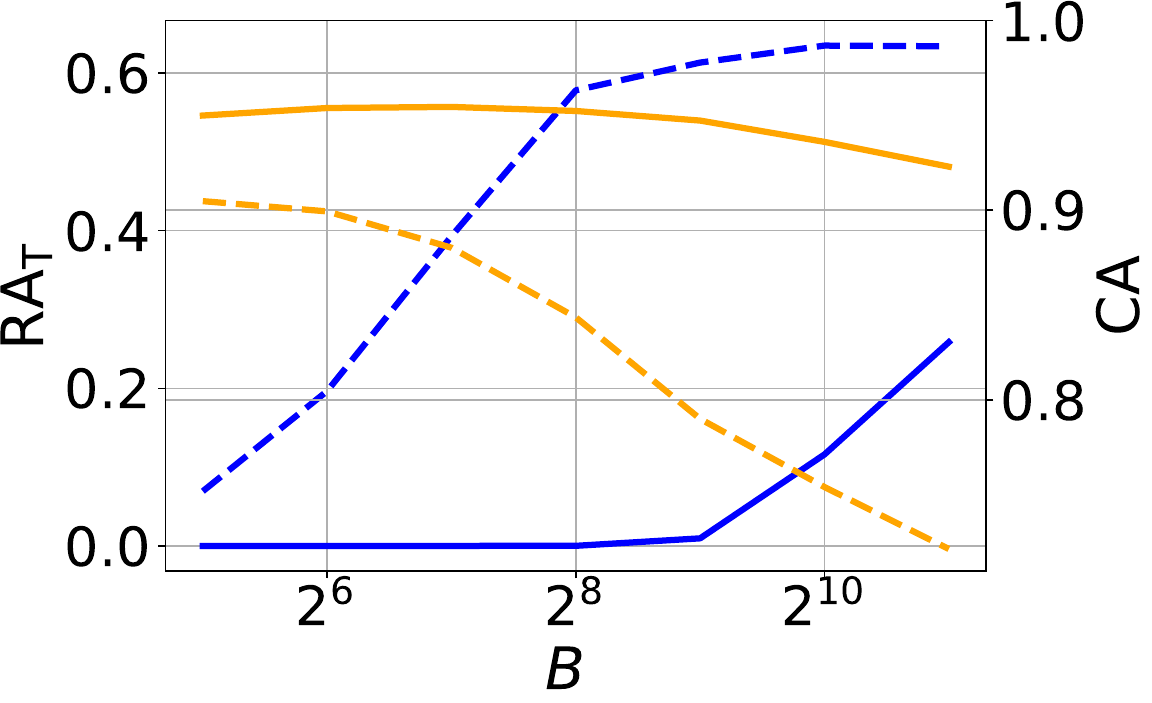}\caption{$\cA_T$ with varying batch size}
\label{fig:transfer_ablation_bs}
\end{subfigure}\\\vspace*{0.5em}
        \begin{subfigure}[t]{.245\textwidth}
        \centering
\includegraphics[width=\linewidth]{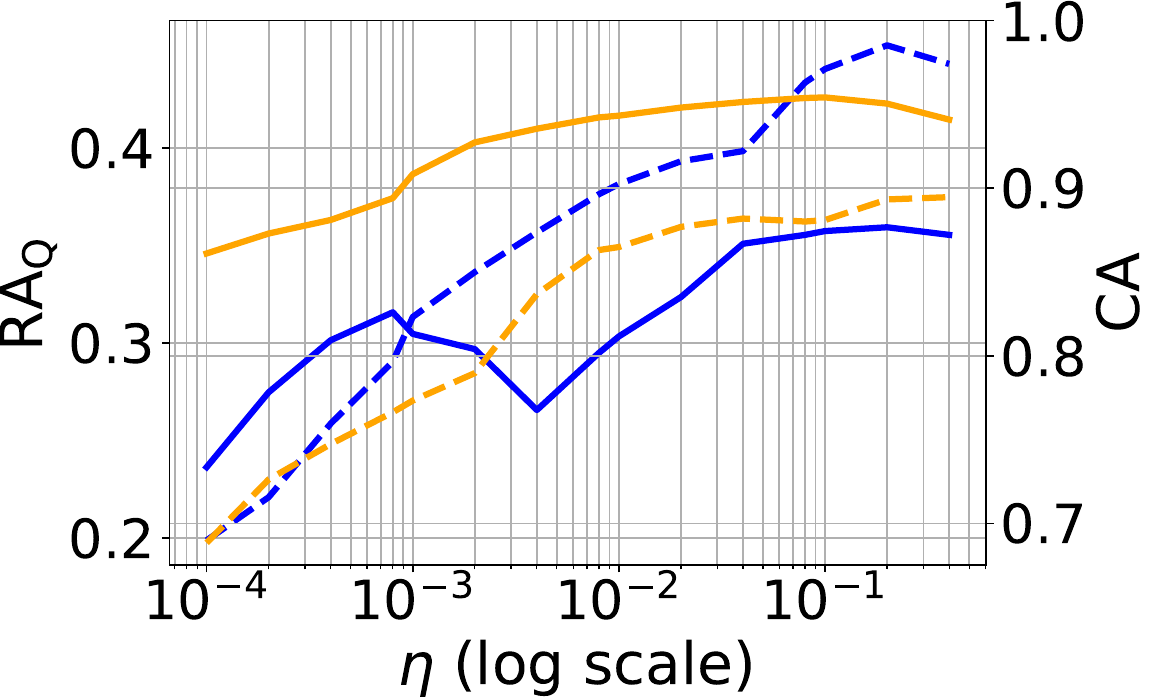}\caption{$\cA_Q$ with varying learning rate}
\label{fig:query_ablation_lr}
\end{subfigure}
        \begin{subfigure}[t]{.245\textwidth}
        \centering
\includegraphics[width=\linewidth]{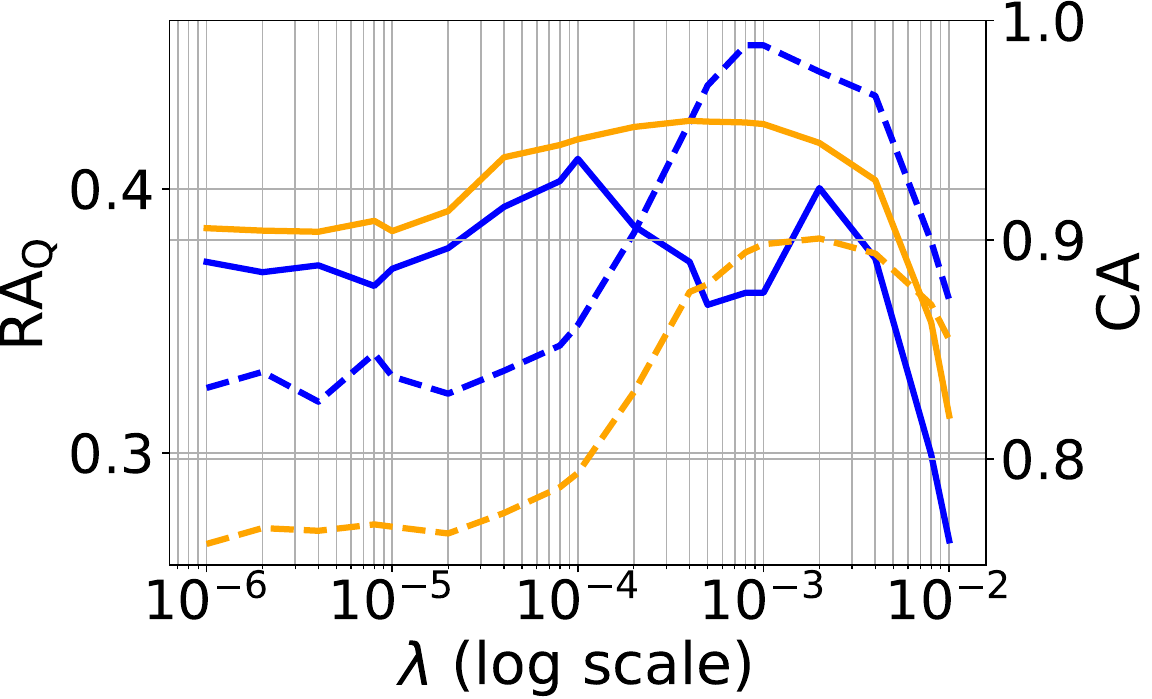}\caption{$\cA_Q$ with varying weight decay}
\label{fig:query_ablation_wd}
\end{subfigure}
        \begin{subfigure}[t]{.245\textwidth}
        \centering
\includegraphics[width=\linewidth]{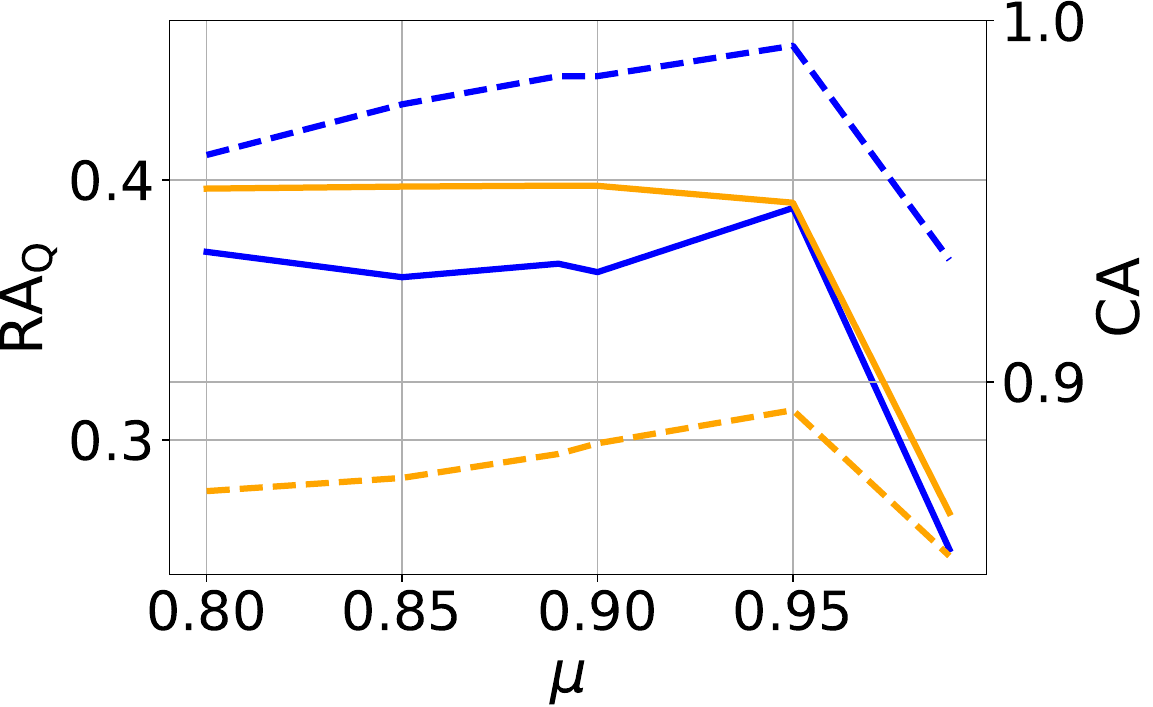}\caption{$\cA_Q$ with varying momentum}
\label{fig:query_ablation_mom}
\end{subfigure}
    \begin{subfigure}[t]{.245\textwidth}
        \centering
\includegraphics[width=\linewidth]{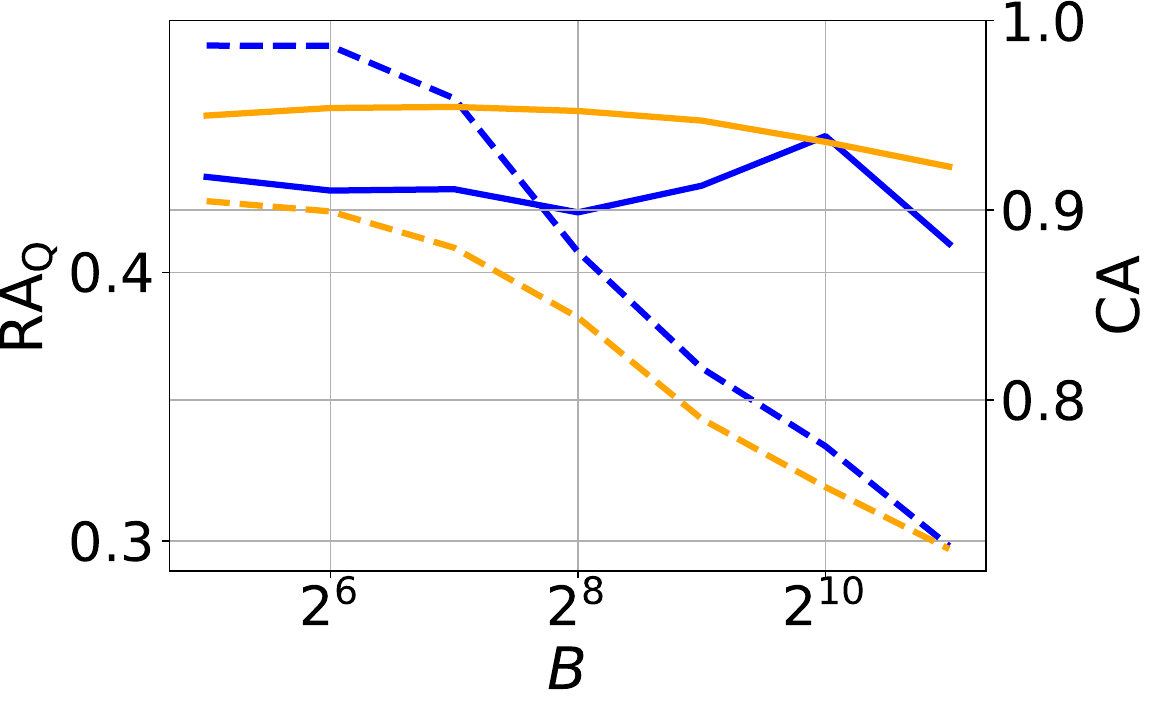}\caption{$\cA_Q$ with varying batch size}
\label{fig:query_ablation_bs}
\end{subfigure}\\
    \caption{Robust accuracy for all hyperparameters $\cH=(\eta, \lambda, \mu, B)$ for transfer-based ($\cA_T$) and query-based attackers ($\cA_Q$). 
    Our results are averaged over all nodes for deep ensembles (solid-line) and distributed ML (dashed-line) on CIFAR-10. The blue line shows \robustacc, while the orange line shows \mainacc. 
    Each column varies the specified hyperparameter, while fixing all others.}
    \label{fig:all_ablations}
\end{figure*}

\vspace{0.25 em}\noindent\textbf{Metrics---(Robust) accuracy.}
Let~$\cS\subset \setX \times \setY$ denote the set of genuine samples provided to the attacker $\adv$,
and let~$\varepsilon$ denote the distortion budget.
To compute the robust accuracy ($\robustacc$), we measure the accuracy on the adversarial examples, i.e., $\dist{x'-x}_p \leq \varepsilon$, generated by the attacker:
\begin{equation}
\robustacc := \nicefrac{\vert
\left\{
  (x,x') \in \mathcal{X}\times\adv(\cS) \;\middle|\;
  C(x)= C(x') 
\right\}\vert}
{\size{\cS}},
\end{equation}
where~$\adv(\cS)$ denotes the set of candidate adversarial examples output by~$\adv$ in a run of the attack on input~$\cS$.
In contrast to the clean accuracy ($\mainacc$) computed on benign images, $\robustacc$ is computed on adversarial examples. The complement of $\robustacc$ is the attack success rate ($\advsuccrate$), i.e., $\advsuccrate = 1 - \robustacc$.

\vspace{0.25 em}\noindent\textbf{Pareto frontier.}
The Pareto frontier emerges as an effective tool to evaluate tradeoffs in multi-objective optimization. In our case, between $\mainacc$ and $\robustacc$. 

A solution~$\omega^*$ is \emph{Pareto optimal} if there exists no other solution that improves all objectives simultaneously.
Formally, given two solutions~$\omega_1$ and~$\omega_2$, we write~$\omega_1 \succ \omega_2$ if~$\omega_1$ dominates~$\omega_2$, i.e., if
$\mainacc(\omega_1) \geq \mainacc(\omega_2) \land \robustacc(\omega_1) \geq \robustacc(\omega_2)$,  where $\mainacc(\omega)$ and~$\robustacc(\omega)$ denote the clean accuracy and robust accuracy as functions of the parameter~$\omega$. The \emph{Pareto frontier} is the set of Pareto-optimal solutions: 
\begin{center}
    $    PF(\Omega) = \{ \omega^*\in \Omega \ | \ \nexists \omega \in \Omega\text{ s.t. } \omega\succ \omega^*\}$.
\end{center}

\section{Experimental Results}

\newcommand{\FullLineComment}[1]{\Statex \texttt{/* #1 */}}
\begin{algorithm}[tb]
\footnotesize
\caption{Ablation of hyperparameters}
\label{alg:hparam_tuning}
\begin{algorithmic}[1]
\State \textbf{Input:} dataset \texttt{data}, ablation ranges  $\texttt{ablation\_ranges}$
\State \textbf{Output:} All ablated models $\texttt{ablated\_models}$\vspace{0.5em}
\FullLineComment{For each type of hyperparameter, we vary it in the given range while keeping the others fixed (cf.~\Cref{tab:parameters}) }
\State ablated\_models = [[], [], [], []]
\State default\_params = [0.1, 0.0005, 0.9, 128]
\For{param\_pos in range(len(['learning\_rate', 'weight\_decay', 'momentum', 'batch\_size']))}
    \State param\_models = []
    \State params = default\_params.copy()
    \For{param\_value in ablation\_ranges[param\_pos]}
        \State params[param\_pos] = param\_value
        \State model = $\mathtt{train\_model}$(params, data)
        \State param\_models.append(model)
    \EndFor
    \State ablated\_models[param\_pos] = param\_models
\EndFor %
\State \textbf{return} ablated\_models
\end{algorithmic}
\end{algorithm}

To answer RQ1 and RQ2, i.e., understand the impact of hyperparameters on the robustness provisions of transfer- ($\rat$) and query-based ($\raq$) attacks, based on the concrete attacks introduced in~\Cref{sec:exps}, respectively, we focus on the four training  hyperparameters that are found in a typical model training using SGD (cf. \Cref{eq:grad_est} and \Cref{eq:theta_update}). The  considered range of hyperparameters can be found in~\Cref{tab:parameters}.
For these, we fix all but one parameter, vary it across a range of values, and monitor the impact on $\mainacc$ and $\robustacc$ of all considered attacks (cf.~\Cref{alg:hparam_tuning}). We average the data points that we obtain across the number of nodes $N$ (each data point is averaged over three independent runs), encompassing centralized deployments ($N=1)$, distributed deployments with ($N=3,5,7$), and data distributions for distributed ML, as deviations originating from these parameters/setups are limited. The impact of $N$ and data distribution is discussed in an ablation study in~\Cref{sec:abl}. 
To answer RQ3, we perform a hyperparameter search (over 100 configurations) using the efficient NSGA-II genetic algorithm~\cite{debFastElitistMultiobjective2002} adapted from Optuna and integrate it in our framework via its Ray integration.

Our default configuration for all experiments consists of a ResNet-18 trained with SGD for 200 epochs (with early stopping) and a CosineAnnealing learning rate scheduler on the CIFAR-10 dataset (with 20\% validation data). 

\vspace{0.25 em}\noindent\textbf{RQ 1:}  
As shown in~\Cref{fig:all_ablations} (first row), we observe a consistent improvement in \robustacc of up to $55\%$ against transfer-based attacks across multiple ML instantiations when the hyperparameters decrease (or batch size $B$ increases). This matches our observations from Equation~\eqref{eq:theta_x} and~\Cref{prop:transfer}.

When considering the impact of the choice of the \textit{learning rate} on deep ensembles, we observe that a reduction of $\eta$ starts to improve \robustacc at $\eta=0.01$, reaching  $\robustacc=0.5$ for $\eta=0.0001$. At the same time, \mainacc maintains an excellent performance of as low as $86\%$. In contrast, the distributed ML instantiations obtain a notably different \mainacc/\robustacc behavior compared to deep ensembles. Here, we see a consistently higher \robustacc across the entire spectrum of $\eta$, already maintaining a $\robustacc=42\%$ at the default $\eta=0.1$ and improving it further to $\robustacc=67\%$ at $\eta=0.001$. Due to the disjoint split data in distributed ML, we see a steeper decline in \mainacc, reaching a still competitive $\mainacc=0.78$ at the highest $\robustacc$. {Concretely, we observe the largest robustness improvement of up to $64\%$ for the i.i.d. distributed ML instance with $N=3$ and $\eta=0.001$.} 

For \textit{weight decay}, we see a similar trend, i.e., an improvement in \robustacc with a decreasing $\lambda$, as this regularizes the model less.
We monitor a \robustacc of only around $27\%$ for deep ensembles, while the distributed ML instantiations reach a $\robustacc=65\%$ at $\mainacc=77\%$ for $\lambda=$ 2e-6 (similar to learning rate).
The \textit{momentum} hyperparameter continues the previous trend, yet improvements for $\robustacc$ for deep ensembles are almost non-existent at a maximum $\robustacc=4\%$. In contrast, distributed ML can reach a $\robustacc=51\%$ and $\mainacc=87\%$ at $\mu=0.8$.
For \textit{batch size}, we start to see an improvement in $\robustacc$ only at $B=2048$ with $\robustacc=26\%$. For distributed ML, we observe a $\robustacc$ as high as $64\%$ at the highest $B$ of 2048. 

\begin{summary}[Robustness against transfer-attacks]
Reducing the value of $\eta, \lambda, \mu$ and increasing $B$ improves \rat with minimal impact on $\mainacc$. Overall, distributed ML  consistently yields higher \rat than deep ensembles trained on the full dataset, with a slight impact on $\mainacc$. This stems from increased ensemble heterogeneity—due to disjoint data and reduced smoothness (as models converge toward sharper minima)—which lowers gradient similarity between the ensemble members (cf. \Cref{prop:transfer}).
\end{summary}

\vspace{0.25 em}\noindent\textbf{RQ 2:} Conforming with our analysis in Equations~\eqref{eq:theta_x} and~\eqref{eq:query}, and in contrast to transfer-based attacks, we observe a consistent impact on \robustacc in~\Cref{fig:all_ablations} with improvements of up to $28\%$ as $\eta, \lambda$ increase and $\mu, B$ decrease. 

We observe that a reduction of $\eta$ reduces \robustacc. For deep ensembles, we obtain the highest $\robustacc=36\%$ at $\eta=0.2$ and see a decline in \robustacc until $\eta=0.004$ at a $\robustacc=27\%$, after which we see a slight increase to $\robustacc=32\%$ at $\eta=0.0008$. Afterwards, \robustacc drops as low as $24\%$. In contrast, we observe a smoother relationship between $\eta$ and $\robustacc$ in distributed ML, achieving the highest $\robustacc=0.46$ at $\eta=0.2$. 

For \textit{weight decay}, we monitor a relatively constant $\robustacc$ for deep ensembles, fluctuating between $36\%$ and $42\%$ across the spectrum of $\lambda$ until $\lambda=0.008$ with a reduction on \mainacc of at most $5\%$. For distributed ML, we obtain a larger range of $\robustacc$ between $32\%$ and $46\%$ at higher fluctuations in $\mainacc$ between $76\%$ and $90\%$. 
For \textit{momentum}, we see a constant \mainacc/\robustacc across the entire range of $\mu$ for both deep ensembles, and distributed ML, with a slight decrease in both \mainacc/\robustacc for $\mu=0.99$. 
For \textit{batch size}, the previously observed trend continues: deep ensembles remain at a consistent \mainacc/\robustacc across the entire hyperparameter range, while distributed ML follows the behavior found with $\eta$, albeit inverted, due to the scaling nature of batch size on learning rate.

\begin{summary}
We find that an increase in $\eta, \lambda$ enhances \raq, while an inverse relationship holds for $\mu, B$. 
Deep ensembles show more stable \raq, likely due to access to the full dataset, leading to flatter minima and smoother input spaces. Distributed ML benefits from this effect especially for larger $\eta$ and $\lambda$.
\end{summary}

\vspace{0.25 em}\noindent\textbf{RQ 3:} Given RQ1 and RQ2, we now investigate whether it is possible to tune hyperparameters to achieve strong robustness against both types of black-box attacks.
To this end, similar to~\cite{lachnitHyperparametersBackdoorResistanceHorizontal2025, bagdasarianMithridatesAuditingBoosting2024}, we make use of the NSGA-II genetic hyperparameter optimization algorithm~\cite{debFastElitistMultiobjective2002} to identify optimal hyperparameter configurations while producing a diverse Pareto front. Specifically, we use the Optuna library with the NSGA-II sampler, integrated into our framework via Ray. Our objective is to find the hyperparameter configuration $\cH$ that simultaneously maximizes both robustness criteria, formalized as follows: $
    \argmax_\cH (\mainacc^\cH, \min(\rat^\cH, \raq^\cH))$.

\begin{figure}
    \centering
    \includegraphics[width=0.95\linewidth]{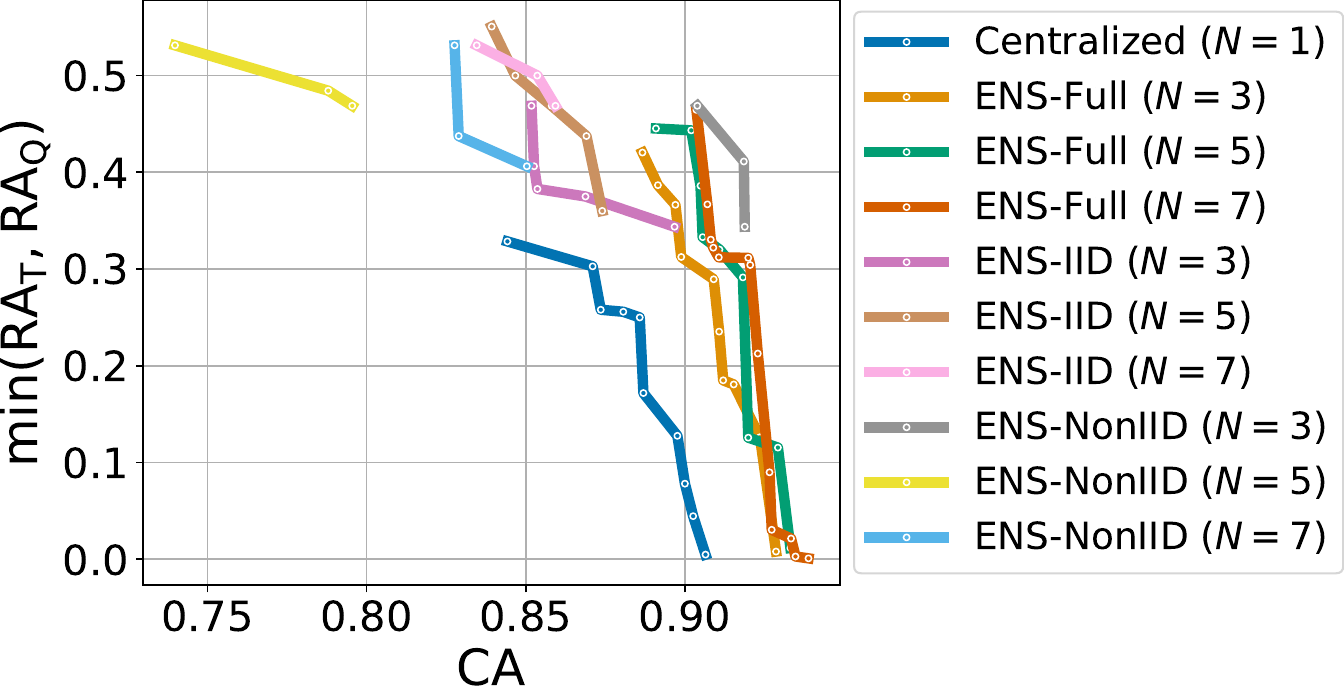}
    \caption{Pareto frontier of the NSGA-II search across all ML instantiations and hyperparameters.}
    \label{fig:nsgaii}
\end{figure}

This approach allows us to identify hyperparameter configurations that (1) maintain strong performance on benign data and (2) guarantee a minimum level of $\robustacc$ against both types of black-box attacks. We run NSGA-II with a population size of 20 and 5 generations, yielding a total of 100 distinct hyperparameter combinations. \Cref{fig:nsgaii} presents the results, illustrating the Pareto fronts for all ML instances. 

We observe the following general trends: deep ensembles significantly enhance both \mainacc and \robustacc as the number of nodes increases, achieving a favorable trade-off at $\mainacc = 0.90$ and $\robustacc = 0.46$ for $N=7$. However, performance gains tend to saturate beyond $N=3$.
In the distributed ML setting with i.i.d. data, results are closely clustered: $N=5$ achieves the highest \robustacc of $55\%$, while $N=3$ yields the best \mainacc at $90\%$. For non-i.i.d. data, the results vary more widely, as the distribution of class splits across nodes significantly impacts performance. Notably, $N=5$ performs the worst in terms of \mainacc, remaining below $80\%$. The best Pareto front is observed at $N=3$.

We now analyze the best-performing configurations for each ML paradigm ($N=7$ for deep ensembles, and $N=5$ and $N=3$ for distributed ML with i.i.d. and non-i.i.d. data, respectively), focusing on \mainacc, \rat, and \raq. We select the best \robustacc values from the hyperparameter search, allowing up to $5\%$ decrease in \mainacc relative to the centralized ($N=1$) baseline. In~\Cref{tab:nsgaii_details}, we compare the results of the best $\cH$ configurations  (see \supmat) to the average \mainacc, \rat, and \raq obtained across all $\cH$ combinations from RQ1 and RQ2 (see \supmat). Overall, we see consistent improvements across all ML paradigms—up to $43\%$ in \rat, $10\%$ in \raq, and $4\%$ in \mainacc.

\vspace{0.25 em}\noindent \textbf{Comparison with SOTA defenses: } Our recommended hyperparameters significantly improve robustness over JPEG~\cite{DBLP:conf/iclr/GuoRCM18} by up to $16\%$ while exhibiting $78\%$ fewer epochs (compared to the default settings in~\Cref{tab:parameters}, reducing from an average of $186$ down to just $41$ epochs) on the same architecture.\footnote{This is due to faster convergence to a well-generalizing minimum, which also leads to earlier onset of overfitting.
} This represents a significant performance gain, especially compared to adversarial training~\cite{rebuffiFixingDataAugmentation2021,gowalUncoveringLimitsAdversarial2021}, which can require up to $40$ GPU hours per model for an average increase of $21\%$ in robustness over our approach.

\begin{summary}
Compared to SOTA defenses like JPEG~\cite{DBLP:conf/iclr/GuoRCM18} and adversarial training~{\cite{gowalUncoveringLimitsAdversarial2021}}, our results in~\Cref{tab:nsgaii_details} demonstrate that our approach (especially in the distributed ML instance) strikes a strong efficiency tradeoff. 
\end{summary}

\begin{table}[tb]
\centering
\begin{adjustbox}{max width=0.73\linewidth}
$\begin{array}{l|l|c|c|c|c|c}
\toprule
 \multirow{3.5}{*}{\textrm{Instance}} &  & \multicolumn{3}{c|}{{\textrm{{Hyperparameter}}}} & \multirow{2}{*}{$\mathsf{JPEG}$} & \multirow{2}{*}{\shortstack{$\mathsf{Adv.}$\\$\mathsf{Training}$}} \\
                             &  &  \multicolumn{3}{c|}{\text{tuning $\cH$}}  &                                    &                                             \\\cmidrule{3-7}
 &  & \text{Avg.} & \text{Ours} & \Delta &  \Delta & \Delta \\\midrule
 \multirow{3}{*}{\textbf{Central.}} & \mainacc & 0.92 & 0.87& -0.05                       &  {-0.02}&{-0.01}\\
                                      & \rat     & 0.08 & 0.51& \mathbf{+0.43}              &                               {-0.16}&                           {+0.18}\\
                                      & \raq     & 0.28 & 0.30& \mathbf{+0.02}              &                               {+0.00}&                           {+0.38}\\
                                      \midrule
\multirow{3.5}{*}{\textbf{ENS-Full}} & \mainacc &  0.94&{0.90}  & {-0.04} &  {-0.03}&{-0.04}\\
                                      & \rat   & 0.07&0.47  & {\mathbf{+0.40}} &                              {-0.10}&                           {+0.22}\\
                                      & \raq   & 0.42&0.46  & {\mathbf{+0.04}}&                               {-0.08}&                           {+0.22}\\
                                      \midrule
\multirow{3.5}{*}{\textbf{ENS-IID}} & \mainacc &  0.86& 0.87  &\mathbf{+0.01}&  {-0.02}&{-0.01}\\
                                      & \rat   &  0.43&0.62  & \mathbf{+0.19}&                               {+0.01}&                           {+0.07}\\
                                      & \raq   &  0.39&0.44  &\mathbf{+0.05}&                               {-0.02}&                           {+0.24}\\
                                      \midrule
\multirow{3.5}{*}{\textbf{\shortstack{ENS-Non-\\IID}}} & \mainacc& 0.86& 0.90& \mathbf{+0.04}& {-0.05}&{-0.04}\\
                                                     & \rat   &  0.32&0.51  & \mathbf{+0.19}&                               {+0.05}&                           {+0.18}\\
                                                     & \raq   &  0.37& 0.47 &\mathbf{+0.10}&                               {-0.07}&                           {+0.21}\\
\bottomrule
\end{array}$
\end{adjustbox}
\caption{Comparison of \mainacc, \robustacc of the recommended $\cH$ found by NSGA-II vs. the default parameters and defenses.}
\label{tab:nsgaii_details}
\end{table}

\section{Ablation Study}
\label{sec:abl}

\begin{table*}[t]
\centering
\begin{adjustbox}{max width=0.75\linewidth}
$\begin{array}{l|c|l|ccccccccccccccc}
\toprule
\multirow{2.5}{*}{\shortstack{Instance}} & \multirow{2.5}{*}{$N$}  & \multirow{2}{*}{} & \multicolumn{15}{c}{\textrm{Learning rate $\eta$}}\\
\cmidrule{4-18}
&&& \textrm{1e-4} & \textrm{2e-4}& \textrm{4e-4} & \textrm{8e-4}& \textrm{1e-3} & \textrm{2e-3} & \textrm{4e-3} & \textrm{8e-3} & 0.01 & 0.02 & 0.04 & 0.08 & 0.1 & 0.2 & 0.4\\
\midrule
 {\multirow{3}{*}{\textbf{Central.}}} & \multirow{3}{*}{1} &\mainacc 
            & 0.83 & 0.84 & 0.85 & 0.87 & 0.89 & 0.91 & 0.92 & 0.93 & 0.93 & 0.94 & 0.94 & 0.95 & 0.95 & 0.94 & 0.93 \\
 &  & \rat  
 & 0.48 & 0.49 & 0.41 & 0.39 & 0.31 & 0.15 & 0.05 & 0.01 & 0.01 & 0.00 & 0.00 & 0.00 & 0.00 & 0.00 & 0.00 \\
 &  & \raq  
 & 0.13 & 0.15 & 0.17 & 0.21 & 0.20 & 0.20 & 0.21 & 0.23 & 0.22 & 0.22 & 0.26 & 0.29 & 0.26 & 0.26 & 0.26 \\
 \midrule

 {\multirow{10}{*}{\textbf{ENS-Full}}}
& \multirow{3}{*}{3} &\mainacc
          & 0.87 & 0.88 & 0.88 & 0.90 & 0.91 & 0.93 & 0.94 & 0.94 & 0.94 & 0.95 & 0.95 & 0.96 & 0.95 & 0.95 & 0.94 \\
& & \rat  
& 0.48 & 0.48 & 0.45 & 0.35 & 0.32 & 0.13 & 0.04 & 0.01 & 0.00 & 0.00 & 0.00 & 0.00 & 0.00 & 0.00 & 0.00 \\
 & &\raq  
 & 0.26 & 0.29 & 0.32 & 0.33 & 0.32 & 0.31 & 0.25 & 0.29 & 0.33 & 0.33 & 0.34 & 0.38 & 0.37 & 0.39 & 0.38 \\\cmidrule{2-18}
& \multirow{3}{*}{5} &\mainacc
         & 0.87 & 0.89 & 0.89 & 0.90 & 0.92 & 0.93 & 0.94 & 0.95 & 0.95 & 0.95 & 0.95 & 0.96 & 0.96 & 0.95 & 0.95 \\
& & \rat  
& 0.51 & 0.51 & 0.47 & 0.37 & 0.28 & 0.12 & 0.04 & 0.01 & 0.01 & 0.00 & 0.00 & 0.00 & 0.00 & 0.00 & 0.00 \\
 & &\raq  
& 0.28 & 0.30 & 0.36 & 0.34 & 0.35 & 0.34 & 0.30 & 0.32 & 0.32 & 0.37 & 0.40 & 0.39 & 0.40 & 0.40 & 0.38 \\\cmidrule{2-18}
& \multirow{3}{*}{7} &\mainacc 
         & 0.87 & 0.88 & 0.89 & 0.90 & 0.92 & 0.93 & 0.94 & 0.95 & 0.95 & 0.95 & 0.95 & 0.96 & 0.96 & 0.95 & 0.95 \\
& & \rat 
& 0.52 & 0.52 & 0.48 & 0.38 & 0.28 & 0.12 & 0.03 & 0.01 & 0.00 & 0.00 & 0.00 & 0.00 & 0.00 & 0.00 & 0.00 \\
 & &\raq 
& 0.28 & 0.35 & 0.35 & 0.39 & 0.35 & 0.34 & 0.31 & 0.34 & 0.35 & 0.37 & 0.41 & 0.37 & 0.40 & 0.39 & 0.41 \\
 \midrule
{\multirow{10}{*}{\textbf{ENS-IID}}} 
& \multirow{3}{*}{3} &\mainacc
          & 0.75 & 0.78 & 0.81 & 0.82 & 0.83 & 0.85 & 0.89 & 0.90 & 0.90 & 0.91 & 0.92 & 0.92 & 0.92 & 0.92 & 0.92 \\
& & \rat  
& 0.62 & 0.65 & 0.66 & 0.64 & 0.66 & 0.63 & 0.49 & 0.38 & 0.30 & 0.20 & 0.10 & 0.12 & 0.07 & 0.02 & 0.02 \\
 & &\raq  
& 0.20 & 0.22 & 0.26 & 0.30 & 0.32 & 0.35 & 0.35 & 0.35 & 0.36 & 0.39 & 0.42 & 0.46 & 0.48 & 0.46 & 0.44 \\\cmidrule{2-18}
& \multirow{3}{*}{5} &\mainacc
          & 0.69 & 0.73 & 0.75 & 0.77 & 0.78 & 0.80 & 0.84 & 0.87 & 0.88 & 0.89 & 0.89 & 0.89 & 0.89 & 0.90 & 0.90 \\
& & \rat  
& 0.62 & 0.65 & 0.67 & 0.68 & 0.69 & 0.68 & 0.66 & 0.59 & 0.57 & 0.48 & 0.39 & 0.48 & 0.45 & 0.14 & 0.06 \\
 & &\raq  
& 0.21 & 0.24 & 0.25 & 0.30 & 0.30 & 0.36 & 0.36 & 0.39 & 0.40 & 0.37 & 0.38 & 0.47 & 0.45 & 0.45 & 0.44 \\\cmidrule{2-18}
& \multirow{3}{*}{7} &\mainacc
          & 0.65 & 0.70 & 0.72 & 0.73 & 0.74 & 0.77 & 0.81 & 0.84 & 0.85 & 0.86 & 0.87 & 0.85 & 0.86 & 0.88 & 0.89 \\
& & \rat  
& 0.61 & 0.63 & 0.65 & 0.67 & 0.67 & 0.70 & 0.70 & 0.67 & 0.67 & 0.62 & 0.57 & 0.63 & 0.63 & 0.47 & 0.17 \\
 & &\raq  
 & 0.20 & 0.23 & 0.28 & 0.30 & 0.29 & 0.31 & 0.38 & 0.39 & 0.39 & 0.42 & 0.42 & 0.39 & 0.41 & 0.46 & 0.48 \\
 \midrule
{\multirow{10}{*}{\textbf{\shortstack{ENS-Non-\\IID}}}}
& \multirow{3}{*}{3} &\mainacc
& 0.73 & 0.76 & 0.78 & 0.80 & 0.81 & 0.81 & 0.87 & 0.88 & 0.87 & 0.88 & 0.89 & 0.90 & 0.90 & 0.90 & 0.90 \\
& & \rat
& 0.61 & 0.62 & 0.63 & 0.63 & 0.63 & 0.61 & 0.48 & 0.45 & 0.42 & 0.34 & 0.21 & 0.22 & 0.21 & 0.03 & 0.02 \\
& &\raq 
& 0.16 & 0.18 & 0.22 & 0.27 & 0.31 & 0.31 & 0.31 & 0.35 & 0.34 & 0.38 & 0.36 & 0.42 & 0.42 & 0.42 & 0.39 \\
\cmidrule{2-18}
& \multirow{3}{*}{5} &\mainacc
& 0.68 & 0.72 & 0.74 & 0.76 & 0.76 & 0.79 & 0.82 & 0.87 & 0.87 & 0.87 & 0.88 & 0.88 & 0.88 & 0.89 & 0.89 \\
& & \rat
& 0.61 & 0.64 & 0.65 & 0.65 & 0.66 & 0.68 & 0.68 & 0.60 & 0.56 & 0.51 & 0.44 & 0.49 & 0.47 & 0.26 & 0.14 \\
 & &\raq
 & 0.21 & 0.23 & 0.27 & 0.30 & 0.33 & 0.36 & 0.36 & 0.39 & 0.40 & 0.39 & 0.40 & 0.46 & 0.48 & 0.46 & 0.42 \\\cmidrule{2-18}
& \multirow{3}{*}{7} &\mainacc
& 0.63 & 0.67 & 0.69 & 0.70 & 0.71 & 0.72 & 0.78 & 0.81 & 0.81 & 0.85 & 0.84 & 0.83 & 0.84 & 0.86 & 0.87 \\
& & \rat
& 0.60 & 0.62 & 0.64 & 0.64 & 0.66 & 0.66 & 0.68 & 0.67 & 0.68 & 0.63 & 0.61 & 0.65 & 0.65 & 0.52 & 0.29 \\
 & &\raq 
& 0.22 & 0.21 & 0.28 & 0.27 & 0.33 & 0.34 & 0.39 & 0.38 & 0.40 & 0.41 & 0.41 & 0.40 & 0.41 & 0.47 & 0.48 \\
\bottomrule
\end{array}$
\end{adjustbox}
\caption{\mainacc and \robustacc of across ML instantiations, and number of nodes using the learning rate hyperparameter $\eta$.}
\label{tab:main_results_lr}
\end{table*}

\begin{table}[t]
\centering
\begin{adjustbox}{max width=0.9\linewidth}
$\begin{array}{l|ccccccccc}
\toprule
 & \multicolumn{9}{c}{\text{Learning rate $\eta$}}\\\cmidrule{2-10}
& \textrm{1e-3} & \textrm{2e-3} & \textrm{4e-3} & \textrm{8e-3} & 0.01 & 0.04 & 0.1 & 0.2 & 0.4\\\midrule
\mainacc & 0.47 & 0.56 & 0.61 & 0.64 & 0.65  & 0.68  & 0.69 & 0.68 & 0.67 \\
\rat& 0.20 & 0.21 & 0.18 & 0.16 & 0.14 &  0.11  & 0.11 & 0.09 & 0.11 \\
\raq& 0.12 & 0.23 & 0.24 & 0.30 & 0.30 &  0.31  & 0.34 & 0.33 & 0.31 \\
\bottomrule
\end{array}$
\end{adjustbox}
\caption{\mainacc and \robustacc on ImageNet using hyperparameter $\eta$.}
\label{tab:ablation_in_adv}
\end{table}

We now ablate our results w.r.t. (1) the number of nodes, (2) data distribution, and (3) complex datasets (e.g., ImageNet). We focus on the learning rate $\eta$, as results for other hyperparameters---provided in the \supmat---show similar trends. 

\vspace{0.25 em}\noindent\textbf{Impact of $\mathbf{N}$.} As shown in~\Cref{tab:main_results_lr}, we observe that higher node counts can achieve up to a $4\%$ improvement in \mainacc, due to their diverse initialization and hence better generalization. While the improvements in \rat are limited to up to $7\%$, \raq improves by up to $20\%$ for the lower learning rates, which we relate to a smoothing of the loss landscape due to the ensembling of diverse, less smooth models.

For distributed ML, we generally observe a steeper decline in \mainacc with an increasing number of nodes, as the total amount of data for each weak learner is reduced. While improvements in \raq are limited to at most $8\%$, \rat strongly improves for $\eta>0.01$, concretely by up to $56\%$ for $\eta=0.1$ in the i.i.d. setup when increasing the node count from 3 to 7. The diversity within the ensemble is increasing with the number of nodes, making it more difficult for adversarial examples to transfer successfully. Both of these results align with our observations while answering RQ3.

\vspace{0.25 em}\noindent\textbf{Impact of Data Distribution.} With respect to the data distribution, we observe that the use of i.i.d. data performs slightly better (by up to $4\%$) in \mainacc than the use of non-i.i.d. data. In general, we see deviations in \rat of up to $12\%$ and \raq of up to $6\%$. In contrast, we observe an interesting behavior when comparing $N=3$ for $\eta>0.01$, where the most significant improvements are seen for non-i.i.d. data. Concretely, we see an improvement in \rat by up to $21\%$, with a decrease in \raq by $8\%$ at a reduction in \mainacc by $3\%$ (confirming our results in RQ3). 

\vspace{0.25 em}\noindent\textbf{Impact of Dataset.} To understand the impact of the dataset on our approach, we adopt our setup used for answering {RQ~1} and {RQ 2}, and train a model on the ImageNet dataset for 90 epochs with a StepLR learning rate scheduler (following the official PyTorch parameters). Due to the larger dataset and longer training times, we focus only on analyzing the impact of the learning rate and vary it across multiple training runs. As shown in~\Cref{tab:ablation_in_adv}, our findings based on CIFAR-10 also hold for ImageNet. Concretely, we see that a reduction in $\eta$ improves \rat, while decreasing \raq.

\section{Conclusion}
In this paper, we demonstrate a striking contrast in how training hyperparameters affect robustness under distinct black-box threat models. Specifically, we find that decreasing the learning rate substantially improves robustness against transfer-based attacks, by up to $64\%$ across various ML instantiations. In contrast, for query-based attacks, we observe that increasing the learning rate improves robustness by up to $28\%$. Supported by thorough experiments, we identify configurations that strike an effective balance---enhancing robustness against both transfer-based and query-based attacks simultaneously. We therefore hope that our findings motivate further research in this fascinating area.

\section*{Acknowledgments}
This work has been co-funded by the Deutsche Forschungsgemeinschaft (DFG, German Research Foundation) under Germany’s Excellence Strategy - EXC 2092 CASA - 390781972 and by the German Federal Ministry of Research, Technology and Space (BMFTR) through the project TRAIN (01IS23027A).

\renewcommand{\thesection}{\Alph{section}}
\setcounter{section}{0}
\section*{Appendix}

\section{Empirical Validation of Main Intuition}
\label{sec:emp_valid}
We empirically validate that (i) implicit regularization through hyperparameters, e.g., learning rate, affects smoothness in the input space, and (ii) this regularization effect reduces gradient similarity. 

For (i), we compute the largest eigenvalues of the input Hessian (which is tractable due to the comparatively low dimensionality of the input space in comparison to the parameter space) for 100 samples of the CIFAR-10 dataset. We empirically estimate the smoothness of a given model $\model$ and data distribution $\cD$ as follows:
    \begin{equation}
        {\sigma}_\model = \bbE_{(x,y)\sim\cD}[\sigma(\nabla_x^2\cL_\model(x,y))]
    \end{equation}
where $\sigma(\cdot)$ denotes the largest eigenvalue, and $\nabla_x^2\cL_\model(x,y)$ the Hessian matrix computed w.r.t. $x$.
Our results are shown in~\Cref{fig:theory_exps} (left) and show that a reduction of learning rate increases the distribution of the largest eigenvalue, indicating that the model becomes less smooth.

For (ii), we use the entire testset of CIFAR-10, i.e., 10,000 samples, for which we evaluate gradients of the input space between a fixed surrogate, trained at $\eta=0.1$, with a target trained on a range of learning rates. To empirically estimate the gradient similarity, we compute it over a given data distribution $\cD$ as: 
    \begin{equation}
        \widetilde{\cS} = \bbE_{(x,y)\sim\cD}[\cS(\cL_\model, \cL_\modelt, x, y)]
    \end{equation}
As shown in~\Cref{fig:theory_exps} (right), the distributions of cosine similarities are wider for higher learning rates, with a median similarity of around $0.25$. Namely, we observe that this distribution becomes narrower with a decrease in the learning rate. Eventually, its median approaches a cosine similarity of $0$, indicating orthogonal gradients, which has shown robustness improvements in other works~\cite{yangTRSTransferabilityReduced}. 

\section{Proof of~\Cref{prop:transfer}}
\label{sec:proof_pop1}
We base our proof on~\cite{yangTRSTransferabilityReduced} and start with their final result on the upper bound on transferability for an untargeted attacker who assumes two models with identical smoothness and bounded gradients. The proof for a targeted attacker is analogous. 

We assume models $\cF$ and $\cG$ are $\beta$-smooth, respectively with bounded gradient magnitude, i.e., $\|\nabla_x \cL(x,y)\| \le B$ for any $x \in \cX$, $y \in \cY$.
            We consider an untargeted attack with perturbation ball $\|\delta\|_2 \le \varepsilon$.
            When the attack radius $\varepsilon$ is small such that the denominator is larger than 0, 
            $\ell_{\min} - \varepsilon B \left(1 + \sqrt{\frac{1+\overline{\cS}(\el,\lt)}{2}}\right) - \beta\epsilon^2 > 0$,
            the transferability can be upper bounded by:
            \resizebox{0.4\textwidth}{!}{
            \begin{minipage}{\linewidth}
            \begin{align*}
                    &\Pr{T_r(\cF, \cG, x) = 1} 
                    \\
                    &\le
                    \dfrac{\xi_\model + \xi_\modelt}{\displaystyle \cL_\mathrm{min} - \varepsilon B \left(1 + \sqrt{\frac{1+\overline{\cS}(\cL_\model,\cL_\modelt)}{2}}\right) - \beta\varepsilon^2}\\
            \end{align*}
            \end{minipage}
            }\\
            Here $\xi_\cF$ and $\xi_\cG$ are the \emph{empirical risks} of models $\cF$ and $\cG$ respectively, defined relative to a differentiable loss. $\cL_\mathrm{min}$ is defined as: $$\min_{\substack{x\in\cX, y'\in\cY: (x,y) \in \supp(\cD), y' \neq y}}(\cL_\model(x, y'), \cL_\modelt(x, y')),$$
            $\supp(\cD)$ is the support of benign data distribution, i.e., $x$ is the benign data and $y$ is its associated true label. 

We now split this upper bound apart and consider the individual properties of each model, i.e., smoothness $\beta_\model, \beta_\modelt$, and gradient magnitudes $\|\nabla_x \cL_\model(x,y)\| \le B_\model$ and $\|\nabla_x \cL_\modelt(x,y)\| \le B_\modelt$:

\resizebox{0.4\textwidth}{!}{
            \begin{minipage}{\linewidth}
            \begin{align*}
                    &\Pr{T_r(\cF, \cG, x) = 1} 
                    \\
                    &\le
                    \dfrac{\xi_\model}{\displaystyle \min_{\substack{x\in\cX, y'\in\cY:\\ (x,y) \in \supp(\cD),\\y' \neq y}}\cL_\model(x, y') - \varepsilon B_\model \left(1 + \sqrt{\frac{1+\overline{\cS}(\cL_\model,\cL_\modelt)}{2}}\right) - \beta_\model\varepsilon^2}\\
                    &+
                    \dfrac{\xi_\modelt}{\displaystyle \min_{\substack{x\in\cX, y'\in\cY:\\ (x,y) \in \supp(\cD),\\y' \neq y}}\cL_\modelt(x, y') - \varepsilon B_\modelt \left(1 + \sqrt{\frac{1+\overline{\cS}(\cL_\model,\cL_\modelt)}{2}}\right) - \beta_\modelt\varepsilon^2}\\
                    &\le
                    \dfrac{\xi_\model + \xi_\modelt}{\displaystyle \cL_\mathrm{min} - \varepsilon B \left(1 + \sqrt{\frac{1+\overline{\cS}(\cL_\model,\cL_\modelt)}{2}}\right) - \beta\varepsilon^2}\\
            \end{align*}
            \end{minipage}
            }\\
In line with~\cite{zhangWhyDoesLittle2024}, we use \Cref{def:smoothness} to locally estimate model smoothness, which is upper-bounded by the global smoothness $\beta_\model$, i.e., $\overline{\sigma}_\model\leq\beta_\model$ and arrive at:

\resizebox{0.4\textwidth}{!}{
            \begin{minipage}{\linewidth}
            \begin{align*}
                    &\Pr{T_r(\cF, \cG, x) = 1} 
                    \\
                    &\le
                    \dfrac{\xi_\model}{\displaystyle \min_{\substack{x\in\cX, y'\in\cY:\\ (x,y) \in \supp(\cD),\\y' \neq y}}\cL_\model(x, y') - \varepsilon B_\model \left(1 + \sqrt{\frac{1+\overline{\cS}(\cL_\model,\cL_\modelt)}{2}}\right) - \overline{\sigma}_\model\varepsilon^2}\\
                    &+
                    \dfrac{\xi_\modelt}{\displaystyle \min_{\substack{x\in\cX, y'\in\cY:\\ (x,y) \in \supp(\cD),\\y' \neq y}}\cL_\modelt(x, y') - \varepsilon B_\modelt \left(1 + \sqrt{\frac{1+\overline{\cS}(\cL_\model,\cL_\modelt)}{2}}\right) - \overline{\sigma}_\modelt\varepsilon^2}\\
            \end{align*}
            \end{minipage}
            }

\section{Results on MobileNetV2}
\label{sec:mobilenet}

We additionally provide experiments on a different model architecture, i.e., MobileNetV2, while varying the learning rate hyperparameter $\eta$ for $N=5$ for non-IID distributed ML. Our results are shown in \Cref{tab:ablation_mn}  for transfer and query-based attacks. These results confirm our main observations, i.e., a reduction in learning rate improves robustness against a transfer-based attacker to up to $64\%$, whereas an increase in learning rate achieves a robustness against a query-based attacker of up to $42\%$.

\begin{table}[htbp]
\centering
\begin{adjustbox}{max width=\linewidth}
$\begin{array}{l|ccccccccccccc}
\toprule
 \multirow{2}{*}{} & \multicolumn{13}{c}{\textrm{Learning rate $\eta$}}\\
\cmidrule{2-14}
& \textrm{1e-4} & \textrm{2e-4}& \textrm{4e-4} & \textrm{8e-4}& \textrm{1e-3} & \textrm{2e-3} & \textrm{4e-3} & \textrm{8e-3} & 0.01 & 0.02 & 0.04 & 0.08 & 0.1\\
\midrule
 \mainacc 
            & 0.46 & 0.57 & 0.66 & 0.73 & 0.74 & 0.74 & 0.79 & 0.79 & 0.81 & 0.84 & 0.88 & 0.90 & 0.88  \\
    \rat  
 & 0.43 & 0.52 & 0.55 & 0.59 & 0.61 & 0.61 & 0.64 & 0.63 & 0.60 & 0.54 & 0.32 & 0.10 & 0.12  \\
    \raq  
& 0.09 & 0.16 & 0.23 & 0.25 & 0.27 & 0.38 & 0.34 & 0.35 & 0.40 & 0.42 & 0.39 & 0.31 & 0.34  \\
\bottomrule
\end{array}$
\end{adjustbox}
\caption{\mainacc and \robustacc on MobileNetV2 using $\eta$.}
\label{tab:ablation_mn}
\end{table}

\section{Full results of RQ1/RQ2}
\label{sec:full_results}
We now provide the full results used to plot~\Cref{fig:all_ablations}. The results for the 
momentum hyperparameter are provided in~\Cref{tab:main_results_mom}, batch size hyperparameter in~\Cref{tab:main_results_bs}, and weight decay hyperparameter in~\Cref{tab:main_results_wd},

\begin{table}[h]
\setlength{\tabcolsep}{4pt}
\centering
\begin{adjustbox}{max height=.67\linewidth}
$\begin{array}{l|c|l|cccccc}
\toprule
\multirow{2.5}{*}{Instance} & \multirow{2.5}{*}{$N$}  & \multirow{2}{*}{} & \multicolumn{6}{c}{\text{Momentum } \mu}\\
\cmidrule{4-9}
&&& 0.8 & 0.85 & 0.89 & 0.9 & 0.95 & 0.99\\
\midrule
\multirow{3}{*}{\textbf{Central.}} & \multirow{3}{*}{1} &\mainacc 
& 0.95 & 0.95 & 0.95 & 0.95 & 0.94 & 0.84 \\
 &  & \rat& 0.00 & 0.00 & 0.00 & 0.00 & 0.00 & 0.04 \\
 &  &\raq & 0.28 & 0.28 & 0.28 & 0.29 & 0.28 & 0.17 \\
 \midrule
\multirow{10}{*}{\textbf{ENS-Full}} 
& \multirow{3}{*}{3} &\mainacc & 0.95 & 0.95 & 0.96 & 0.96 & 0.95 & 0.87 \\
& & \rat& 0.00 & 0.00 & 0.00 & 0.00 & 0.00 & 0.03 \\
 & &\raq & 0.38 & 0.36 & 0.38 & 0.38 & 0.38 & 0.27 \\\cmidrule{2-9}
& \multirow{3}{*}{5} &\mainacc & 0.96 & 0.96 & 0.96 & 0.96 & 0.95 & 0.87 \\ 
& & \rat& 0.00 & 0.00 & 0.00 & 0.00 & 0.00 & 0.04 \\
 & &\raq & 0.42 & 0.40 & 0.40 & 0.40 & 0.44 & 0.29 \\\cmidrule{2-9}
& \multirow{3}{*}{7} &\mainacc& 0.96 & 0.96 & 0.96 & 0.96 & 0.95 & 0.88 \\
& & \rat& 0.00 & 0.00 & 0.00 & 0.00 & 0.00 & 0.03 \\
 & &\raq & 0.41 & 0.41 & 0.41 & 0.40 & 0.46 & 0.30 \\
 \midrule
\multirow{10}{*}{\textbf{ENS-IID}} 
& \multirow{3}{*}{3} &\mainacc & 0.91 & 0.92 & 0.92 & 0.92 & 0.92 & 0.88 \\
& & \rat& 0.24 & 0.15 & 0.08 & 0.07 & 0.01 & 0.04 \\
 & &\raq& 0.44 & 0.46 & 0.45 & 0.45 & 0.44 & 0.32 \\\cmidrule{2-9}
& \multirow{3}{*}{5} &\mainacc& 0.89 & 0.89 & 0.90 & 0.89 & 0.90 & 0.87 \\
& & \rat& 0.52 & 0.54 & 0.42 & 0.38 & 0.19 & 0.07 \\
 & &\raq & 0.41 & 0.44 & 0.45 & 0.45 & 0.46 & 0.39 \\\cmidrule{2-9}
& \multirow{3}{*}{7} &\mainacc& 0.85 & 0.85 & 0.85 & 0.86 & 0.88 & 0.85 \\
& & \rat& 0.64 & 0.66 & 0.63 & 0.65 & 0.52 & 0.16 \\
 & &\raq & 0.38 & 0.42 & 0.40 & 0.42 & 0.46 & 0.43 \\
 \midrule
\multirow{10}{*}{\textbf{\shortstack{ENS-Non-\\IID}}} 
& \multirow{3}{*}{3} &\mainacc & 0.89 & 0.89 & 0.90 & 0.90 & 0.90 & 0.85 \\
& & \rat& 0.39 & 0.35 & 0.19 & 0.19 & 0.04 & 0.11 \\
 & &\raq & 0.41 & 0.44 & 0.45 & 0.43 & 0.43 & 0.30 \\\cmidrule{2-9}
& \multirow{3}{*}{5} &\mainacc & 0.87 & 0.87 & 0.88 & 0.88 & 0.89 & 0.85 \\
& & \rat& 0.58 & 0.54 & 0.50 & 0.49 & 0.26 & 0.16 \\
 & &\raq& 0.42 & 0.42 & 0.48 & 0.47 & 0.47 & 0.37 \\\cmidrule{2-9}
& \multirow{3}{*}{7} &\mainacc & 0.82 & 0.83 & 0.83 & 0.85 & 0.86 & 0.82 \\
& & \rat& 0.66 & 0.65 & 0.65 & 0.61 & 0.50 & 0.31 \\
 & &\raq & 0.39 & 0.39 & 0.41 & 0.42 & 0.46 & 0.41 \\
\bottomrule
\end{array}$
\end{adjustbox}
\caption{\mainacc and \robustacc of across ML instantiations, and number of nodes using the momentum hyperparameter $\mu$.}
\label{tab:main_results_mom}
\end{table}

\begin{table}[t]
\setlength{\tabcolsep}{4pt}
\centering
\begin{adjustbox}{max height=.6\linewidth}
$\begin{array}{l|c|l|ccccccccccccccc}
\toprule
\multirow{2.5}{*}{Instance} & \multirow{2.5}{*}{$N$}  & \multirow{2}{*}{} & \multicolumn{7}{c}{\text{Batch size } B}\\
\cmidrule{4-10}
&&& 32 & 64 & 128 & 256 & 512 & 1024 & 2048\\
\midrule
\multirow{3}{*}{\textbf{Central.}} & \multirow{3}{*}{1} &\mainacc 
& 0.94 & 0.95 & 0.95 & 0.95 & 0.94 & 0.93 & 0.91 \\
&  & \rat& 0.00 & 0.00 & 0.00 & 0.00 & 0.02 & 0.12 & 0.27 \\
 &  &\raq & 0.36 & 0.36 & 0.34 & 0.32 & 0.33 & 0.36 & 0.28 \\
 \midrule
\multirow{10}{*}{\textbf{ENS-Full}} 
& \multirow{3}{*}{3} &\mainacc & 0.95 & 0.95 & 0.96 & 0.95 & 0.95 & 0.94 & 0.93 \\
& & \rat& 0.00 & 0.00 & 0.00 & 0.00 & 0.01 & 0.11 & 0.23 \\
 & &\raq & 0.42 & 0.42 & 0.42 & 0.43 & 0.43 & 0.46 & 0.42 \\\cmidrule{2-10}
& \multirow{3}{*}{5} &\mainacc & 0.95 & 0.96 & 0.96 & 0.96 & 0.95 & 0.94 & 0.93 \\
& & \rat& 0.00 & 0.00 & 0.00 & 0.00 & 0.00 & 0.13 & 0.27 \\
 & &\raq & 0.49 & 0.46 & 0.48 & 0.46 & 0.47 & 0.48 & 0.46 \\\cmidrule{2-10}
& \multirow{3}{*}{7} &\mainacc & 0.95 & 0.96 & 0.96 & 0.96 & 0.95 & 0.94 & 0.93 \\
& & \rat& 0.00 & 0.00 & 0.00 & 0.00 & 0.01 & 0.11 & 0.27 \\
 & &\raq & 0.48 & 0.48 & 0.48 & 0.47 & 0.49 & 0.50 & 0.47 \\
 \midrule
\multirow{10}{*}{\textbf{ENS-IID}} 
& \multirow{3}{*}{3} &\mainacc & 0.93 & 0.93 & 0.92 & 0.91 & 0.89 & 0.85 & 0.79 \\
& & \rat& 0.01 & 0.01 & 0.05 & 0.44 & 0.51 & 0.62 & 0.66 \\
 & &\raq & 0.48 & 0.48 & 0.48 & 0.45 & 0.41 & 0.36 & 0.34 \\\cmidrule{2-10}
& \multirow{3}{*}{5} &\mainacc& 0.91 & 0.91 & 0.89 & 0.87 & 0.79 & 0.75 & 0.74 \\
& & \rat& 0.03 & 0.11 & 0.39 & 0.61 & 0.66 & 0.65 & 0.65 \\
 & &\raq & 0.48 & 0.50 & 0.48 & 0.42 & 0.36 & 0.35 & 0.31 \\\cmidrule{2-10}
& \multirow{3}{*}{7} &\mainacc& 0.90 & 0.89 & 0.85 & 0.80 & 0.73 & 0.71 & 0.69 \\
& & \rat& 0.11 & 0.38 & 0.66 & 0.69 & 0.67 & 0.65 & 0.65 \\
 & &\raq & 0.51 & 0.49 & 0.44 & 0.38 & 0.35 & 0.32 & 0.28 \\
 \midrule
\multirow{10}{*}{\textbf{\shortstack{ENS-Non-\\IID}}} 
& \multirow{3}{*}{3} &\mainacc & 0.91 & 0.91 & 0.90 & 0.87 & 0.84 & 0.82 & 0.76 \\
& & \rat& 0.01 & 0.04 & 0.14 & 0.44 & 0.56 & 0.60 & 0.60 \\
 & &\raq & 0.44 & 0.43 & 0.45 & 0.39 & 0.35 & 0.34 & 0.30 \\\cmidrule{2-10}
& \multirow{3}{*}{5} &\mainacc & 0.90 & 0.89 & 0.88 & 0.84 & 0.81 & 0.73 & 0.71 \\
& & \rat& 0.06 & 0.17 & 0.48 & 0.63 & 0.66 & 0.66 & 0.63 \\
 & &\raq& 0.50 & 0.51 & 0.50 & 0.43 & 0.38 & 0.33 & 0.31 \\\cmidrule{2-10}
& \multirow{3}{*}{7} &\mainacc & 0.89 & 0.87 & 0.84 & 0.77 & 0.68 & 0.66 & 0.64 \\
& & \rat& 0.20 & 0.47 & 0.64 & 0.66 & 0.63 & 0.63 & 0.61 \\
 & &\raq & 0.49 & 0.50 & 0.44 & 0.37 & 0.33 & 0.30 & 0.26 \\
\bottomrule
\end{array}$
\end{adjustbox}
\caption{\mainacc and \robustacc of across ML instantiations, and number of nodes using the batch size hyperparameter $B$.}
\label{tab:main_results_bs}
\end{table}

\begin{table*}[tbp]
\footnotesize
\setlength{\tabcolsep}{4pt}
\centering
\begin{adjustbox}{max width=0.98\linewidth}
$\begin{array}{l|c|l|cccccccccccccccccc}
\toprule
\multirow{2.5}{*}{\text{Instance}} & \multirow{2.5}{*}{$N$}  & \multirow{2}{*}{} & \multicolumn{18}{c}{\text{Weight decay } \lambda}\\
\cmidrule{4-21}
&&& \text{1e-6} & \text{2e-6}& \text{4e-6} & \text{8e-6}& \text{1e-5} & \text{2e-5} & \text{4e-5} & \text{8e-5} & \text{1e-4} & \text{2e-4} & \text{4e-4} & \text{5e-4} & \text{8e-4} & \text{1e-3} & \text{2e-3} & \text{4e-3}& \text{8e-3}& \text{1e-2}\\
\midrule
\multirow{3}{*}{\textbf{Central.}} & \multirow{3}{*}{1} &\mainacc 
& 0.88 & 0.88 & 0.87 & 0.89 & 0.88 & 0.89 & 0.93 & 0.94 & 0.94 & 0.94 & 0.95 & 0.95 & 0.95 & 0.95 & 0.93 & 0.91 & 0.84 & 0.79 \\
 &  & \rat
& 0.16 & 0.27 & 0.41 & 0.18 & 0.18 & 0.17 & 0.05 & 0.01 & 0.01 & 0.00 & 0.00 & 0.00 & 0.00 & 0.00 & 0.00 & 0.01 & 0.08 & 0.11 \\
 &  &\raq 
& 0.27 & 0.28 & 0.26 & 0.23 & 0.25 & 0.29 & 0.30 & 0.34 & 0.32 & 0.32 & 0.30 & 0.28 & 0.28 & 0.29 & 0.33 & 0.27 & 0.16 & 0.16 \\
 \midrule
\multirow{10}{*}{\textbf{ENS-Full}} 
& \multirow{3}{*}{3} &\mainacc & 0.91 & 0.90 & 0.91 & 0.91 & 0.91 & 0.91 & 0.94 & 0.94 & 0.95 & 0.95 & 0.96 & 0.95 & 0.95 & 0.95 & 0.95 & 0.93 & 0.86 & 0.83 \\
& & \rat& 0.23 & 0.33 & 0.22 & 0.17 & 0.14 & 0.19 & 0.06 & 0.02 & 0.01 & 0.00 & 0.00 & 0.00 & 0.00 & 0.00 & 0.00 & 0.00 & 0.03 & 0.11 \\
 & &\raq& 0.38 & 0.39 & 0.38 & 0.39 & 0.37 & 0.41 & 0.39 & 0.40 & 0.43 & 0.35 & 0.37 & 0.36 & 0.37 & 0.35 & 0.38 & 0.38 & 0.31 & 0.25 \\\cmidrule{2-21}
& \multirow{3}{*}{5} &\mainacc & 0.91 & 0.92 & 0.92 & 0.92 & 0.91 & 0.92 & 0.94 & 0.95 & 0.95 & 0.95 & 0.96 & 0.96 & 0.96 & 0.96 & 0.95 & 0.93 & 0.87 & 0.83 \\
& & \rat& 0.21 & 0.16 & 0.19 & 0.16 & 0.25 & 0.16 & 0.03 & 0.01 & 0.00 & 0.00 & 0.00 & 0.00 & 0.00 & 0.00 & 0.00 & 0.00 & 0.03 & 0.12 \\
 & &\raq & 0.40 & 0.39 & 0.40 & 0.41 & 0.41 & 0.39 & 0.44 & 0.41 & 0.45 & 0.44 & 0.42 & 0.40 & 0.38 & 0.39 & 0.44 & 0.42 & 0.34 & 0.32 \\\cmidrule{2-21}
& \multirow{3}{*}{7} &\mainacc & 0.91 & 0.91 & 0.92 & 0.92 & 0.92 & 0.92 & 0.94 & 0.95 & 0.95 & 0.96 & 0.96 & 0.96 & 0.96 & 0.96 & 0.95 & 0.93 & 0.87 & 0.83 \\
& & \rat& 0.25 & 0.29 & 0.22 & 0.25 & 0.23 & 0.17 & 0.04 & 0.02 & 0.00 & 0.00 & 0.00 & 0.00 & 0.00 & 0.00 & 0.00 & 0.00 & 0.03 & 0.11 \\
 & &\raq & 0.43 & 0.42 & 0.45 & 0.42 & 0.45 & 0.41 & 0.44 & 0.47 & 0.45 & 0.44 & 0.40 & 0.38 & 0.41 & 0.41 & 0.46 & 0.42 & 0.38 & 0.34 \\
 \midrule
\multirow{10}{*}{\textbf{ENS-IID}} 
& \multirow{3}{*}{3} &\mainacc & 0.82 & 0.84 & 0.83 & 0.84 & 0.84 & 0.83 & 0.83 & 0.84 & 0.89 & 0.91 & 0.92 & 0.92 & 0.93 & 0.92 & 0.92 & 0.91 & 0.88 & 0.86 \\
& & \rat& 0.64 & 0.62 & 0.62 & 0.60 & 0.62 & 0.63 & 0.59 & 0.60 & 0.53 & 0.35 & 0.11 & 0.06 & 0.01 & 0.01 & 0.01 & 0.01 & 0.04 & 0.07 \\
 & &\raq& 0.35 & 0.36 & 0.30 & 0.34 & 0.35 & 0.31 & 0.32 & 0.36 & 0.41 & 0.44 & 0.44 & 0.46 & 0.44 & 0.43 & 0.45 & 0.42 & 0.32 & 0.32 \\\cmidrule{2-21}
& \multirow{3}{*}{5} &\mainacc& 0.77 & 0.78 & 0.78 & 0.80 & 0.78 & 0.78 & 0.79 & 0.79 & 0.78 & 0.86 & 0.89 & 0.89 & 0.90 & 0.91 & 0.91 & 0.90 & 0.88 & 0.87 \\
& & \rat& 0.65 & 0.67 & 0.65 & 0.63 & 0.65 & 0.65 & 0.65 & 0.65 & 0.67 & 0.64 & 0.51 & 0.48 & 0.24 & 0.12 & 0.03 & 0.03 & 0.07 & 0.10 \\
 & &\raq & 0.31 & 0.32 & 0.32 & 0.33 & 0.30 & 0.34 & 0.33 & 0.34 & 0.34 & 0.37 & 0.45 & 0.45 & 0.47 & 0.47 & 0.47 & 0.46 & 0.39 & 0.35 \\\cmidrule{2-21}
& \multirow{3}{*}{7} &\mainacc & 0.74 & 0.74 & 0.73 & 0.74 & 0.73 & 0.74 & 0.74 & 0.74 & 0.75 & 0.78 & 0.85 & 0.85 & 0.88 & 0.88 & 0.89 & 0.89 & 0.88 & 0.87 \\
& & \rat& 0.67 & 0.66 & 0.66 & 0.67 & 0.67 & 0.65 & 0.67 & 0.66 & 0.66 & 0.67 & 0.65 & 0.63 & 0.55 & 0.47 & 0.14 & 0.06 & 0.08 & 0.13 \\
 & &\raq & 0.34 & 0.34 & 0.33 & 0.34 & 0.33 & 0.32 & 0.36 & 0.33 & 0.33 & 0.35 & 0.42 & 0.40 & 0.46 & 0.48 & 0.47 & 0.48 & 0.44 & 0.39 \\
 \midrule
\multirow{10}{*}{\textbf{\shortstack{ENS-Non-\\IID}}} 
& \multirow{3}{*}{3} &\mainacc & 0.79 & 0.80 & 0.81 & 0.79 & 0.79 & 0.80 & 0.82 & 0.87 & 0.87 & 0.87 & 0.90 & 0.90 & 0.91 & 0.91 & 0.90 & 0.89 & 0.85 & 0.83 \\
& & \rat& 0.60 & 0.62 & 0.57 & 0.61 & 0.62 & 0.60 & 0.54 & 0.47 & 0.50 & 0.43 & 0.23 & 0.18 & 0.05 & 0.04 & 0.02 & 0.03 & 0.10 & 0.14 \\
 & &\raq & 0.29 & 0.29 & 0.30 & 0.32 & 0.30 & 0.32 & 0.31 & 0.32 & 0.30 & 0.39 & 0.41 & 0.44 & 0.42 & 0.41 & 0.38 & 0.37 & 0.33 & 0.30 \\\cmidrule{2-21}
& \multirow{3}{*}{5} &\mainacc & 0.76 & 0.76 & 0.78 & 0.77 & 0.77 & 0.76 & 0.78 & 0.78 & 0.78 & 0.85 & 0.87 & 0.88 & 0.89 & 0.90 & 0.90 & 0.89 & 0.87 & 0.86 \\
& & \rat& 0.66 & 0.65 & 0.64 & 0.66 & 0.66 & 0.66 & 0.66 & 0.65 & 0.65 & 0.63 & 0.58 & 0.52 & 0.33 & 0.19 & 0.07 & 0.04 & 0.10 & 0.17 \\
 & &\raq& 0.34 & 0.34 & 0.34 & 0.37 & 0.35 & 0.33 & 0.36 & 0.34 & 0.33 & 0.42 & 0.45 & 0.48 & 0.47 & 0.48 & 0.44 & 0.42 & 0.39 & 0.38 \\\cmidrule{2-21}
& \multirow{3}{*}{7} &\mainacc & 0.69 & 0.69 & 0.69 & 0.69 & 0.69 & 0.69 & 0.69 & 0.70 & 0.70 & 0.72 & 0.83 & 0.84 & 0.86 & 0.87 & 0.88 & 0.88 & 0.86 & 0.85 \\
& & \rat& 0.64 & 0.64 & 0.64 & 0.64 & 0.64 & 0.63 & 0.64 & 0.65 & 0.63 & 0.66 & 0.67 & 0.64 & 0.57 & 0.51 & 0.25 & 0.14 & 0.19 & 0.28 \\
 & &\raq & 0.31 & 0.34 & 0.34 & 0.33 & 0.34 & 0.32 & 0.32 & 0.34 & 0.37 & 0.33 & 0.38 & 0.40 & 0.45 & 0.46 & 0.45 & 0.46 & 0.42 & 0.40 \\
\bottomrule
\end{array}$
\end{adjustbox}
\caption{\mainacc and \robustacc of across ML instantiations, and number of nodes using the weight decay hyperparameter $\lambda$.}
\label{tab:main_results_wd}
\end{table*}

\section{Recommended Hyperparameters}
\label{sec:full_hparams}
In Table~\ref{tab:best_hparams}, we list the recommended hyperparameters $\cH$ output by NSGA-II when analyzing RQ3.
\begin{table*}[t]
\setlength{\tabcolsep}{4pt}
\centering
\begin{adjustbox}{max width=\linewidth}
$\begin{array}{l|c|l|l|l|l}
\toprule
\text{Instance} & N & \text{Learning rate } \eta& \text{Weight decay } \lambda & \text{Momentum } \mu & \text{Batch size } B\\\midrule
 \textbf{Central.} & 1&0.128513963751756 & 0.00002106648601704221 & 0.8711945357302524& 1024\\\midrule
\textbf{ENS-Full}& 7&0.14466065922350002 & 0.0001117820920069873& 0.8033175657855712& 1024\\\midrule
\textbf{ENS-IID} & 5&0.0016925498065751279& 0.004997749370303174&0.8226947042481179 &32\\\midrule
\textbf{ENS-Non-IID} & 3&0.05222863774453028& 0.0000025852079056761314& 0.9136617206670944&32\\
\bottomrule
\end{array}$
\end{adjustbox}
\caption{Recommended choice of $\cH$ output by NSGA-II.}
\label{tab:best_hparams}
\end{table*}


\begin{thebibliography}{40}
\providecommand{\natexlab}[1]{#1}

\bibitem[{Andreina, Zimmer, and Karame(2025)}]{DBLP:conf/aaai/AndreinaZK25}
Andreina, S.; Zimmer, P.; and Karame, G. 2025.
\newblock On the Robustness of Distributed Machine Learning against Transfer Attacks.
\newblock In Walsh, T.; Shah, J.; and Kolter, Z., eds., \emph{{{AAAI-25}}, Sponsored by the Association for the Advancement of Artificial Intelligence, February 25 - March 4, 2025, Philadelphia, {{PA}}, {{USA}}}, 15382--15390. AAAI Press.

\bibitem[{Andriushchenko et~al.(2020)Andriushchenko, Croce, Flammarion, and Hein}]{10.1007/978-3-030-58592-1_29}
Andriushchenko, M.; Croce, F.; Flammarion, N.; and Hein, M. 2020.
\newblock Square Attack: A Query-Efficient Black-Box Adversarial Attack via Random Search.
\newblock In \emph{Computer Vision -- {{ECCV}} 2020: 16th European Conference, Glasgow, {{UK}}, August 23--28, 2020, Proceedings, Part {{XXIII}}}, 484--501. Berlin, Heidelberg: Springer-Verlag.
\newblock ISBN 978-3-030-58591-4.

\bibitem[{Bagdasarian and Shmatikov(2024)}]{bagdasarianMithridatesAuditingBoosting2024}
Bagdasarian, E.; and Shmatikov, V. 2024.
\newblock Mithridates: {{Auditing}} and {{Boosting Backdoor Resistance}} of {{Machine Learning Pipelines}}.
\newblock In \emph{Proceedings of the 2024 on {{ACM SIGSAC Conference}} on {{Computer}} and {{Communications Security}}}, 4480--4494. Salt Lake City UT USA: ACM.
\newblock ISBN 979-8-4007-0636-3.

\bibitem[{Biggio et~al.(2013)Biggio, Corona, Maiorca, Nelson, Srndic, Laskov, Giacinto, and Roli}]{DBLP:conf/pkdd/BiggioCMNSLGR13}
Biggio, B.; Corona, I.; Maiorca, D.; Nelson, B.; Srndic, N.; Laskov, P.; Giacinto, G.; and Roli, F. 2013.
\newblock Evasion Attacks against Machine Learning at Test Time.
\newblock In Blockeel, H.; Kersting, K.; Nijssen, S.; and Zelezn{\'y}, F., eds., \emph{Machine Learning and Knowledge Discovery in Databases - European Conference, {{ECML PKDD}} 2013, Prague, Czech Republic, September 23-27, 2013, Proceedings, Part {{III}}}, volume 8190 of \emph{Lecture Notes in Computer Science}, 387--402. Springer.

\bibitem[{Cai et~al.(2023)Cai, Ning, Yang, and Wang}]{caiEnsembleinOneEnsembleLearning2023}
Cai, Y.; Ning, X.; Yang, H.; and Wang, Y. 2023.
\newblock Ensemble-in-{{One}}: {{Ensemble Learning}} within {{Random Gated Networks}} for {{Enhanced Adversarial Robustness}}.
\newblock \emph{Proceedings of the AAAI Conference on Artificial Intelligence}, 37(12): 14738--14747.

\bibitem[{Chen et~al.(2024)Chen, Zhang, Dong, Yang, Su, and Zhu}]{chenRethinkingModelEnsemble2024}
Chen, H.; Zhang, Y.; Dong, Y.; Yang, X.; Su, H.; and Zhu, J. 2024.
\newblock Rethinking Model Ensemble in Transfer-Based Adversarial Attacks.
\newblock In \emph{The Twelfth International Conference on Learning Representations, {{ICLR}} 2024, Vienna, Austria, May 7-11, 2024}. OpenReview.net.

\bibitem[{Chen, Jordan, and Wainwright(2020)}]{chenHopSkipJumpAttackQueryEfficientDecisionBased2020}
Chen, J.; Jordan, M.~I.; and Wainwright, M.~J. 2020.
\newblock {{HopSkipJumpAttack}}: {{A Query-Efficient Decision-Based Attack}}.
\newblock In \emph{2020 {{IEEE Symposium}} on {{Security}} and {{Privacy}} ({{SP}})}, 1277--1294. San Francisco, CA, USA: IEEE.
\newblock ISBN 978-1-7281-3497-0.

\bibitem[{Chen et~al.(2017)Chen, Zhang, Sharma, Yi, and Hsieh}]{chenZOOZerothOrder2017}
Chen, P.-Y.; Zhang, H.; Sharma, Y.; Yi, J.; and Hsieh, C.-J. 2017.
\newblock {{ZOO}}: {{Zeroth Order Optimization Based Black-box Attacks}} to {{Deep Neural Networks}} without {{Training Substitute Models}}.
\newblock In \emph{Proceedings of the 10th {{ACM Workshop}} on {{Artificial Intelligence}} and {{Security}}}, 15--26. Dallas Texas USA: ACM.
\newblock ISBN 978-1-4503-5202-4.

\bibitem[{Cohen et~al.(2021)Cohen, Kaur, Li, Kolter, and Talwalkar}]{cohenGradientDescentNeural2022}
Cohen, J.; Kaur, S.; Li, Y.; Kolter, J.~Z.; and Talwalkar, A. 2021.
\newblock Gradient Descent on Neural Networks Typically Occurs at the Edge of Stability.
\newblock In \emph{9th International Conference on Learning Representations, {{ICLR}} 2021, Virtual Event, Austria, May 3-7, 2021}. OpenReview.net.

\bibitem[{Croce and Hein(2020)}]{DBLP:conf/icml/Croce020a}
Croce, F.; and Hein, M. 2020.
\newblock Reliable Evaluation of Adversarial Robustness with an Ensemble of Diverse Parameter-Free Attacks.
\newblock In \emph{Proceedings of the 37th International Conference on Machine Learning, {{ICML}} 2020, 13-18 July 2020, Virtual Event}, volume 119 of \emph{Proceedings of Machine Learning Research}, 2206--2216. PMLR.

\bibitem[{D'Angelo et~al.(2024)D'Angelo, Andriushchenko, Varre, and Flammarion}]{dangeloWhyWeNeed}
D'Angelo, F.; Andriushchenko, M.; Varre, A.~V.; and Flammarion, N. 2024.
\newblock Why Do We Need Weight Decay in Modern Deep Learning?
\newblock In Globersons, A.; Mackey, L.; Belgrave, D.; Fan, A.; Paquet, U.; Tomczak, J.~M.; and Zhang, C., eds., \emph{Advances in Neural Information Processing Systems 38: {{Annual}} Conference on Neural Information Processing Systems 2024, {{NeurIPS}} 2024, Vancouver, {{BC}}, Canada, December 10 - 15, 2024}.

\bibitem[{Deb et~al.(2002)Deb, Pratap, Agarwal, and Meyarivan}]{debFastElitistMultiobjective2002}
Deb, K.; Pratap, A.; Agarwal, S.; and Meyarivan, T. 2002.
\newblock A Fast and Elitist Multiobjective Genetic Algorithm: {{NSGA-II}}.
\newblock \emph{IEEE Transactions on Evolutionary Computation}, 6(2): 182--197.

\bibitem[{Demontis et~al.(2019)Demontis, Melis, Pintor, Jagielski, Biggio, Oprea, {Nita-Rotaru}, and Roli}]{demontis2019adversarial}
Demontis, A.; Melis, M.; Pintor, M.; Jagielski, M.; Biggio, B.; Oprea, A.; {Nita-Rotaru}, C.; and Roli, F. 2019.
\newblock Why Do Adversarial Attacks Transfer? {{Explaining}} Transferability of Evasion and Poisoning Attacks.
\newblock In \emph{28th {{USENIX}} Security Symposium ({{USENIX}} Security 19)}, 321--338.

\bibitem[{Deng and Mu(2023)}]{dengUnderstandingImprovingEnsemble}
Deng, Y.; and Mu, T. 2023.
\newblock Understanding and Improving Ensemble Adversarial Defense.
\newblock In Oh, A.; Naumann, T.; Globerson, A.; Saenko, K.; Hardt, M.; and Levine, S., eds., \emph{Advances in Neural Information Processing Systems 36: {{Annual}} Conference on Neural Information Processing Systems 2023, {{NeurIPS}} 2023, New Orleans, {{LA}}, {{USA}}, December 10 - 16, 2023}.

\bibitem[{Dherin et~al.(2022)Dherin, Munn, Rosca, and Barrett}]{dherinWhyNeuralNetworksa}
Dherin, B.; Munn, M.; Rosca, M.; and Barrett, D. 2022.
\newblock Why Neural Networks Find Simple Solutions: {{The}} Many Regularizers of Geometric Complexity.
\newblock In Koyejo, S.; Mohamed, S.; Agarwal, A.; Belgrave, D.; Cho, K.; and Oh, A., eds., \emph{Advances in Neural Information Processing Systems 35: {{Annual}} Conference on Neural Information Processing Systems 2022, {{NeurIPS}} 2022, New Orleans, {{LA}}, {{USA}}, November 28 - December 9, 2022}.

\bibitem[{Dong et~al.(2018)Dong, Liao, Pang, Su, Zhu, Hu, and Li}]{DBLP:conf/cvpr/DongLPS0HL18}
Dong, Y.; Liao, F.; Pang, T.; Su, H.; Zhu, J.; Hu, X.; and Li, J. 2018.
\newblock Boosting Adversarial Attacks with Momentum.
\newblock In \emph{2018 {{IEEE}} Conference on Computer Vision and Pattern Recognition, {{CVPR}} 2018, Salt Lake City, {{UT}}, {{USA}}, June 18-22, 2018}, 9185--9193. Computer Vision Foundation / IEEE Computer Society.

\bibitem[{Goodfellow, Shlens, and Szegedy(2015)}]{goodfellowExplainingHarnessingAdversarial2015}
Goodfellow, I.~J.; Shlens, J.; and Szegedy, C. 2015.
\newblock Explaining and Harnessing Adversarial Examples.
\newblock In Bengio, Y.; and LeCun, Y., eds., \emph{3rd International Conference on Learning Representations, {{ICLR}} 2015, San Diego, {{CA}}, {{USA}}, May 7-9, 2015, Conference Track Proceedings}.

\bibitem[{Gowal et~al.(2021)Gowal, Qin, Uesato, Mann, and Kohli}]{gowalUncoveringLimitsAdversarial2021}
Gowal, S.; Qin, C.; Uesato, J.; Mann, T.; and Kohli, P. 2021.
\newblock Uncovering the {{Limits}} of {{Adversarial Training}} against {{Norm-Bounded Adversarial Examples}}.
\newblock arXiv:2010.03593.

\bibitem[{Guo et~al.(2018)Guo, Rana, Ciss{\'e}, and {van der Maaten}}]{DBLP:conf/iclr/GuoRCM18}
Guo, C.; Rana, M.; Ciss{\'e}, M.; and {van der Maaten}, L. 2018.
\newblock Countering Adversarial Images Using Input Transformations.
\newblock In \emph{6th International Conference on Learning Representations, {{ICLR}} 2018, Vancouver, {{BC}}, Canada, April 30 - May 3, 2018, Conference Track Proceedings}. OpenReview.net.

\bibitem[{He et~al.(2016)He, Zhang, Ren, and Sun}]{heDeepResidualLearning2016}
He, K.; Zhang, X.; Ren, S.; and Sun, J. 2016.
\newblock Deep Residual Learning for Image Recognition.
\newblock In \emph{2016 {{IEEE}} Conference on Computer Vision and Pattern Recognition, {{CVPR}} 2016, Las Vegas, {{NV}}, {{USA}}, June 27-30, 2016}, 770--778. IEEE Computer Society.

\bibitem[{Ilyas et~al.(2018)Ilyas, Engstrom, Athalye, and Lin}]{ilyasBlackboxAdversarialAttacksa}
Ilyas, A.; Engstrom, L.; Athalye, A.; and Lin, J. 2018.
\newblock Black-Box Adversarial Attacks with Limited Queries and Information.
\newblock In Dy, J.~G.; and Krause, A., eds., \emph{Proceedings of the 35th International Conference on Machine Learning, {{ICML}} 2018, Stockholmsm{\"a}ssan, Stockholm, Sweden, July 10-15, 2018}, volume~80 of \emph{Proceedings of Machine Learning Research}, 2142--2151. PMLR.

\bibitem[{Jastrz{\k e}bski et~al.(2018)Jastrz{\k e}bski, Kenton, Arpit, Ballas, Fischer, Bengio, and Storkey}]{jastrzebskiThreeFactorsInfluencing2018}
Jastrz{\k e}bski, S.; Kenton, Z.; Arpit, D.; Ballas, N.; Fischer, A.; Bengio, Y.; and Storkey, A. 2018.
\newblock Three {{Factors Influencing Minima}} in {{SGD}}.
\newblock arXiv:1711.04623.

\bibitem[{Kariyappa and Qureshi(2019)}]{kariyappaImprovingAdversarialRobustness2019}
Kariyappa, S.; and Qureshi, M.~K. 2019.
\newblock Improving {{Adversarial Robustness}} of {{Ensembles}} with {{Diversity Training}}.
\newblock arXiv:1901.09981.

\bibitem[{Kaur, Cohen, and Lipton(2022)}]{kaurMaximumHessianEigenvalue}
Kaur, S.; Cohen, J.; and Lipton, Z.~C. 2022.
\newblock On the Maximum Hessian Eigenvalue and Generalization.
\newblock In Antor{\'a}n, J.; Blaas, A.; Feng, F.; Ghalebikesabi, S.; Mason, I.; Pradier, M.~F.; Rohde, D.; Ruiz, F. J.~R.; and Schein, A., eds., \emph{Proceedings on "{{I}} Can't Believe It's Not Better! - {{Understanding}} Deep Learning through Empirical Falsification" at {{NeurIPS}} 2022 Workshops, 03 December 2022, New Orleans, Louisiana, {{USA}}}, volume 187 of \emph{Proceedings of Machine Learning Research}, 51--65. PMLR.

\bibitem[{Krizhevsky(2009)}]{cifar10}
Krizhevsky, A. 2009.
\newblock Learning Multiple Layers of Features from Tiny Images.
\newblock Technical report.

\bibitem[{Kuang et~al.(2024)Kuang, Liu, Lin, and Ji}]{kuangDefenseAdversarialAttacks2024}
Kuang, H.; Liu, H.; Lin, X.; and Ji, R. 2024.
\newblock Defense {Against} {Adversarial} {Attacks} {Using} {Topology} {Aligning} {Adversarial} {Training}.
\newblock \emph{IEEE Transactions on Information Forensics and Security}, 19: 3659--3673.

\bibitem[{Lachnit and Karame(2025)}]{lachnitHyperparametersBackdoorResistanceHorizontal2025}
Lachnit, S.; and Karame, G. 2025.
\newblock On Hyperparameters and Backdoor-Resistance in Horizontal Federated Learning.
\newblock In \emph{Proceedings of the 2025 ACM SIGSAC Conference on Computer and Communications Security}, CCS '25, 1919–1933. New York, NY, USA: Association for Computing Machinery.
\newblock ISBN 9798400715259.

\bibitem[{Ma and Ying(2021)}]{maLinearStabilitySGD}
Ma, C.; and Ying, L. 2021.
\newblock On Linear Stability of {{SGD}} and Input-Smoothness of Neural Networks.
\newblock In Ranzato, M.; Beygelzimer, A.; Dauphin, Y.; Liang, P.; and Vaughan, J.~W., eds., \emph{Advances in Neural Information Processing Systems}, volume~34, 16805--16817. Curran Associates, Inc.

\bibitem[{{Moosavi-Dezfooli} et~al.(2019){Moosavi-Dezfooli}, Fawzi, Uesato, and Frossard}]{moosavi-dezfooliRobustnessCurvatureRegularization2019}
{Moosavi-Dezfooli}, S.-M.; Fawzi, A.; Uesato, J.; and Frossard, P. 2019.
\newblock Robustness via {{Curvature Regularization}}, and {{Vice Versa}}.
\newblock In \emph{2019 {{IEEE}}/{{CVF Conference}} on {{Computer Vision}} and {{Pattern Recognition}} ({{CVPR}})}, 9070--9078. Long Beach, CA, USA: IEEE.
\newblock ISBN 978-1-7281-3293-8.

\bibitem[{Nie et~al.(2022)Nie, Guo, Huang, Xiao, Vahdat, and Anandkumar}]{DBLP:conf/icml/NieGHXVA22}
Nie, W.; Guo, B.; Huang, Y.; Xiao, C.; Vahdat, A.; and Anandkumar, A. 2022.
\newblock Diffusion Models for Adversarial Purification.
\newblock In Chaudhuri, K.; Jegelka, S.; Song, L.; Szepesv{\'a}ri, C.; Niu, G.; and Sabato, S., eds., \emph{International Conference on Machine Learning, {{ICML}} 2022, 17-23 July 2022, Baltimore, Maryland, {{USA}}}, volume 162 of \emph{Proceedings of Machine Learning Research}, 16805--16827. PMLR.

\bibitem[{Pang et~al.(2019)Pang, Xu, Du, Chen, and Zhu}]{pangImprovingAdversarialRobustness}
Pang, T.; Xu, K.; Du, C.; Chen, N.; and Zhu, J. 2019.
\newblock Improving Adversarial Robustness via Promoting Ensemble Diversity.
\newblock In Chaudhuri, K.; and Salakhutdinov, R., eds., \emph{Proceedings of the 36th International Conference on Machine Learning, {{ICML}} 2019, 9-15 June 2019, Long Beach, California, {{USA}}}, volume~97 of \emph{Proceedings of Machine Learning Research}, 4970--4979. PMLR.

\bibitem[{Rebuffi et~al.(2021)Rebuffi, Gowal, Calian, Stimberg, Wiles, and Mann}]{rebuffiFixingDataAugmentation2021}
Rebuffi, S.-A.; Gowal, S.; Calian, D.~A.; Stimberg, F.; Wiles, O.; and Mann, T. 2021.
\newblock Fixing {{Data Augmentation}} to {{Improve Adversarial Robustness}}.
\newblock arXiv:2103.01946.

\bibitem[{Russakovsky et~al.(2015)Russakovsky, Deng, Su, Krause, Satheesh, Ma, Huang, Karpathy, Khosla, Bernstein, Berg, and {Fei-Fei}}]{russakovskyImageNetLargeScale2015}
Russakovsky, O.; Deng, J.; Su, H.; Krause, J.; Satheesh, S.; Ma, S.; Huang, Z.; Karpathy, A.; Khosla, A.; Bernstein, M.; Berg, A.~C.; and {Fei-Fei}, L. 2015.
\newblock {{ImageNet Large Scale Visual Recognition Challenge}}.
\newblock \emph{International Journal of Computer Vision}, 115(3): 211--252.

\bibitem[{Szegedy et~al.(2014)Szegedy, Zaremba, Sutskever, Bruna, Erhan, Goodfellow, and Fergus}]{DBLP:journals/corr/SzegedyZSBEGF13}
Szegedy, C.; Zaremba, W.; Sutskever, I.; Bruna, J.; Erhan, D.; Goodfellow, I.~J.; and Fergus, R. 2014.
\newblock Intriguing Properties of Neural Networks.
\newblock In Bengio, Y.; and LeCun, Y., eds., \emph{2nd International Conference on Learning Representations, {{ICLR}} 2014, Banff, {{AB}}, Canada, April 14-16, 2014, Conference Track Proceedings}.

\bibitem[{Tram{\`e}r et~al.(2018)Tram{\`e}r, Kurakin, Papernot, Goodfellow, Boneh, and McDaniel}]{DBLP:conf/iclr/TramerKPGBM18}
Tram{\`e}r, F.; Kurakin, A.; Papernot, N.; Goodfellow, I.~J.; Boneh, D.; and McDaniel, P.~D. 2018.
\newblock Ensemble Adversarial Training: {{Attacks}} and Defenses.
\newblock In \emph{6th International Conference on Learning Representations, {{ICLR}} 2018, Vancouver, {{BC}}, Canada, April 30 - May 3, 2018, Conference Track Proceedings}. OpenReview.net.

\bibitem[{Tu et~al.(2019)Tu, Ting, Chen, Liu, Zhang, Yi, Hsieh, and Cheng}]{tuAutoZOOMAutoencoderBasedZeroth2019}
Tu, C.-C.; Ting, P.; Chen, P.-Y.; Liu, S.; Zhang, H.; Yi, J.; Hsieh, C.-J.; and Cheng, S.-M. 2019.
\newblock {{AutoZOOM}}: {{Autoencoder-Based Zeroth Order Optimization Method}} for {{Attacking Black-Box Neural Networks}}.
\newblock \emph{Proceedings of the AAAI Conference on Artificial Intelligence}, 33: 742--749.

\bibitem[{Wu et~al.(2021)Wu, Su, Lyu, and King}]{DBLP:conf/cvpr/WuSLK21}
Wu, W.; Su, Y.; Lyu, M.~R.; and King, I. 2021.
\newblock Improving the Transferability of Adversarial Samples with Adversarial Transformations.
\newblock In \emph{{{IEEE}} Conference on Computer Vision and Pattern Recognition, {{CVPR}} 2021, Virtual, June 19-25, 2021}, 9024--9033. Computer Vision Foundation / IEEE.

\bibitem[{Yang et~al.(2020)Yang, Zhang, Dong, Inkawhich, Gardner, Touchet, Wilkes, Berry, and Li}]{yangDVERGEDiversifyingVulnerabilities}
Yang, H.; Zhang, J.; Dong, H.; Inkawhich, N.; Gardner, A.; Touchet, A.; Wilkes, W.; Berry, H.; and Li, H. 2020.
\newblock {{DVERGE}}: Diversifying Vulnerabilities for Enhanced Robust Generation of Ensembles.
\newblock In Larochelle, H.; Ranzato, M.; Hadsell, R.; Balcan, M.-F.; and Lin, H.-T., eds., \emph{Advances in Neural Information Processing Systems 33: {{Annual}} Conference on Neural Information Processing Systems 2020, {{NeurIPS}} 2020, December 6-12, 2020, Virtual}.

\bibitem[{Yang et~al.(2021)Yang, Li, Xu, Zuo, Chen, Zhou, Rubinstein, Zhang, and Li}]{yangTRSTransferabilityReduced}
Yang, Z.; Li, L.; Xu, X.; Zuo, S.; Chen, Q.; Zhou, P.; Rubinstein, B. I.~P.; Zhang, C.; and Li, B. 2021.
\newblock {{TRS}}: Transferability Reduced Ensemble via Promoting Gradient Diversity and Model Smoothness.
\newblock In Ranzato, M.; Beygelzimer, A.; Dauphin, Y.~N.; Liang, P.; and Vaughan, J.~W., eds., \emph{Advances in Neural Information Processing Systems 34: {{Annual}} Conference on Neural Information Processing Systems 2021, {{NeurIPS}} 2021, December 6-14, 2021, Virtual}, 17642--17655.

\bibitem[{Zhang et~al.(2024)Zhang, Hu, Zhang, Shi, Li, Liu, Wan, and Jin}]{zhangWhyDoesLittle2024}
Zhang, Y.; Hu, S.; Zhang, L.~Y.; Shi, J.; Li, M.; Liu, X.; Wan, W.; and Jin, H. 2024.
\newblock Why {{Does Little Robustness Help}}? {{A Further Step Towards Understanding Adversarial Transferability}}.
\newblock In \emph{2024 {{IEEE Symposium}} on {{Security}} and {{Privacy}} ({{SP}})}, 3365--3384. San Francisco, CA, USA: IEEE.
\newblock ISBN 979-8-3503-3130-1.

\end{thebibliography}
\end{document}